\documentclass[10pt,journal,compsoc]{IEEEtran}
\ifCLASSINFOpdf
\else
\fi

\usepackage{hyperref}
\usepackage{graphicx}
\usepackage{url}
\usepackage{paralist,tabularx}
\usepackage{amsfonts}
\usepackage[dvipsnames]{xcolor}
\usepackage{color}
\usepackage{enumitem}
\usepackage{makecell}
\usepackage{multirow}
\usepackage{transparent}
\usepackage{booktabs}

\definecolor{barblue}{cmyk}{0.89,0.66,0.31,0}
\newlength\MAX  
\newlength\MAXX  
\newcommand*\Chart[1]{\rlap{#1\%}\transparent{0.3}\textcolor{barblue}{\rule{#1\MAXX}{2ex}}}

\newlength\NAA 
\newcommand*\textChart[1]{\rlap{#1}\transparent{0.3}\textcolor{barblue}{\rule{#1\NAA}{2ex}}}

\begin{document}
\title{VisImages: A Fine-Grained Expert-Annotated Visualization Dataset}
% \title{VisImages: A Multi-Level Visualization Dataset from VIS Publications}

\author{Dazhen Deng, Yihong Wu, Xinhuan Shu, Jiang Wu, Siwei Fu, Weiwei Cui, Yingcai Wu% <-this % stops a space
\IEEEcompsocitemizethanks{\IEEEcompsocthanksitem D. Deng, Y. Wu, J. Wu, and Y. Wu were with the State Key Lab of CAD\&CG, Zhejiang University, Hangzhou, China.\protect\\
E-mail: \{dengdazhen, wuyihong, wujiang5521, ycwu\}@zju.edu.cn
\IEEEcompsocthanksitem X. Shu was with Department of Computer Science and Engineering, the Hong Kong University of Science and Technology, Hong Kong.\protect\\
E-mail: xinhuan.shu@connect.ust.hk
\IEEEcompsocthanksitem S. Fu was with Zhejiang Lab, Hangzhou, China.\protect\\
E-mail: fusiwei339@gmail.com
\IEEEcompsocthanksitem W. Cui was with Microsoft Research Asia, Beijing, China.\protect\\
E-mail: \{weiweicu\}@microsoft.com
\IEEEcompsocthanksitem Yingcai Wu is the corresponding author.}

% \thanks{Manuscript received 8 June 2021; revised 1 Feb. 2022; accepted 14 Feb. 2022.}
}


\markboth{IEEE TRANSACTIONS ON VISUALIZATION AND COMPUTER GRAPHICS}%
{Shell \MakeLowercase{\textit{et al.}}: Bare Demo of IEEEtran.cls for Computer Society Journals}



\IEEEtitleabstractindextext{%
\begin{abstract}
Images in visualization publications contain rich information, e.g., novel visualization designs and implicit design patterns of visualizations. A systematic collection of these images can contribute to the community in many aspects, such as literature analysis and automated tasks for visualization. In this paper, we build and make public a dataset, VisImages, which collects 12,267 images with captions from 1,397 papers in IEEE InfoVis and VAST. Built upon a comprehensive visualization taxonomy, the dataset includes 35,096 visualizations and their bounding boxes in the images.
We demonstrate the usefulness of VisImages through three use cases: 1) investigating the use of visualizations in the publications with VisImages Explorer, 2) training and benchmarking models for visualization classification, and 3) localizing visualizations in the visual analytics systems automatically.

\end{abstract}

}

\maketitle

\IEEEdisplaynontitleabstractindextext

% \fi
\IEEEpeerreviewmaketitle

\IEEEraisesectionheading{\section{Introduction}\label{sec:introduction}}
% \IEEEPARstart{A}s the saying goes, a picture is worth a thousand words.
\IEEEPARstart{I}mages are crucial to publications in the visualization community (e.g., IEEE VIS), showcasing the visual designs, system frameworks, model details, experiment results, etc. 
The images contain a rich trove of visual information (e.g., color schemes and shapes of graphical elements) and semantic information (e.g., different combinations of charts) that can advance the understanding of the field and the research of artificial intelligence for visualization (AI4VIS)~\cite{wu2021survey}.

First, the information contained in the images can greatly benefit the analysis of visualization literature, which mainly employs publication metadata like keywords, citations, and co-authorship~\cite{federico2016survey}.
Incorporating image data into literature analysis can understand the visualization field from more dimensions (e.g., what types of charts are frequently used in different venues across years; how different charts are used in different research topics?) and inspire new research problems (e.g., how different charts can be organized together to better represent data?).

Moreover, a visualization dataset from visualization publications affords new opportunities for the application of AI4VIS.
Existing studies~\cite{battle2018beagle,savva2011revision,siegel2016figureseer} collect chart images online and train computer vision models for visualization tasks, such as chart classification.
The images in these datasets usually comprise common chart types with simple layouts due to the data source.
Therefore, the models trained on these datasets might fail when dealing with charts with complex designs, such as system interfaces with multiple charts.
Consequently, a new visualization dataset from visualization publications can advance the research in developing machine learning models and serve as a benchmark to test the generalizability and robustness of models.
\begin{figure}[tb]
    \centering
    \includegraphics[width=\linewidth]{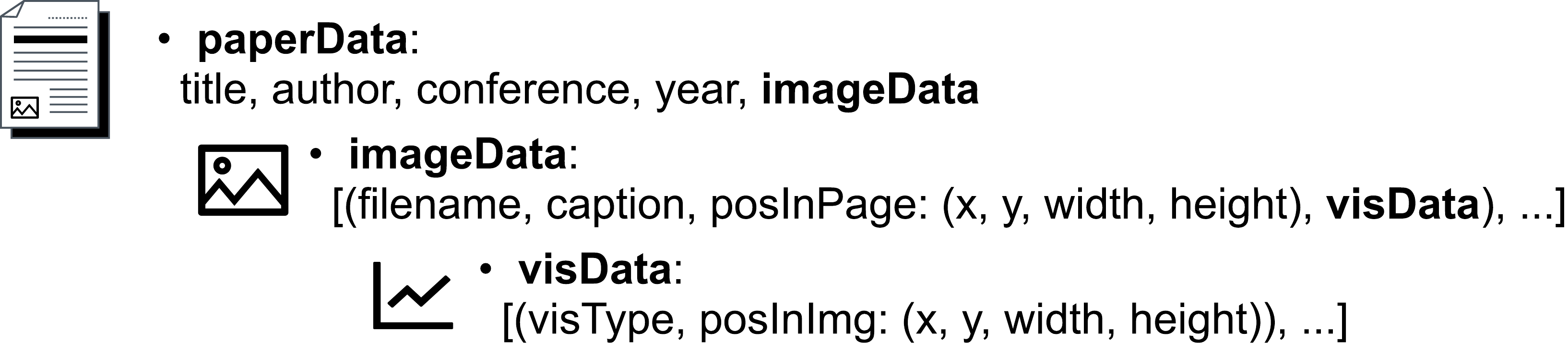}
    \caption{Data structure of VisImages.}
    \label{fig:data-structure}
\end{figure}

However, the images from visualization publications cannot be directly utilized for above tasks, as they lack adequate annotations that describe the semantics information in the images. 
% However, there is a gap to fill when utilizing the images from visualization publications for literature analysis or training machine learning models.
% The gap is adequate annotations that describe the semantic information in the images.
For example, when analyzing visualization literature, the information of the visualization types in the images is necessary to index and search the images of interest for in-depth analysis.
In addition, there is a lack of proper labels for applying frontier machine learning models to visualization tasks (e.g., deconstructing visual analytics systems or generating visualizations).
In particular, object detection models require the bounding boxes of visualizations in the images, and image-text translation models need textual descriptions about the images.
\begin{figure*}[!htb]
    \centering
    \includegraphics[width=\linewidth]{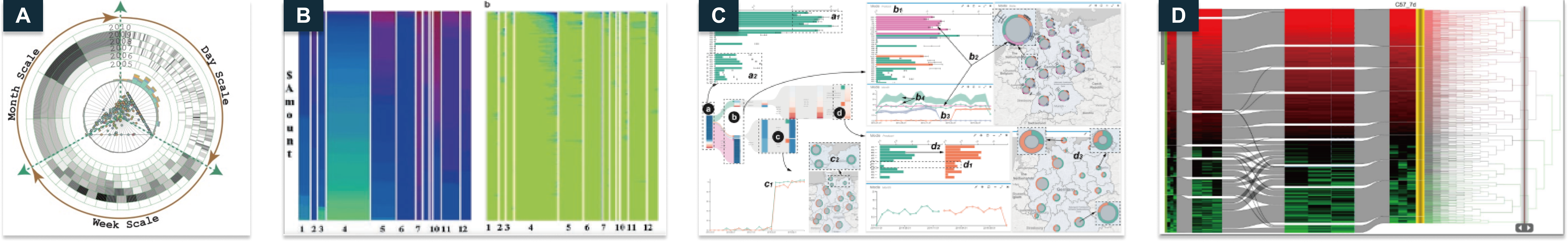}
    \caption{Novel designs in visualization academic publications. (A) shows the design of OpinionSeer~\cite{wu2010opinionseer}, consisting of a triangular scatterplot, a circular bar chart, and donut charts. (B) shows the pixel bar charts~\cite{Keim2002PixelBC}, a variant of bar chart. (C) shows the interface of TPFlow~\cite{liu2018tpflow}, which is facilitated with a set of views of different visualizations. (D) shows a design by Lex et al.~\cite{lex2010comparative}, which combines a Sankey diagram, heatmaps, and a tree.}
    \label{fig:challenge}
\end{figure*}
 
% Fortunately, in visualization publications, the figures are naturally accompanied with captions, which have the potential to fill this gap with appropriate annotation.

To facilitate the use of images, we build and make public a visualization dataset, \textbf{VisImages}, from visualization publications.
The data in VisImages is organized into three levels, namely, papers, images, and visualizations (\figurename~\ref{fig:data-structure}). The paper data includes metadata of the paper (i.e., title, authors, conference, and year) and image data. The metadata of the paper is coded from vispubdata.org~\cite{isenberg2016vispubdata}. The image data is a list of data for each image, which includes the image file name, textual caption, image position (i.e., bounding box) in the paper, and visualization data.
The visualization data is a list of data for each visualization, including the visualization type and visualization position (i.e., bounding box) in the image.
In all, the dataset contains the data of 1,397 papers, 12,267 images, and 35,096 visualizations.

% with comprehensive samples and rich annotations.
% We attempt to curate and examine the visualizations crafted by the academic community itself, which have a gulf of knowledge from visualization researchers.
% With the dataset, we hope to open up a wide range of significant applications such as literature analysis and automated tasks in the visualization community.

Creating such a dataset faces three major challenges.~\label{sec:challenge} 
The first challenge is categorizing diverse visualizations.
Though various taxonomies have been proposed to define and distinguish different visualization types~\cite{shneiderman1996eyes, borkin2013makes}, they cannot fully cover the various designs in visualization publications, such as novel glyphs (\figurename\ref{fig:challenge}-A) or the variations of existing visualizations (\figurename\ref{fig:challenge}-B).
In addition, annotating visualization types requires extensive visualization expertise.
Second, the diverse layouts (\figurename\ref{fig:challenge}-C\&D) and large quantity of visualizations in the publications makes identifying the positions of visualizations difficult and tedious.
% For example, the images include single charts (\figurename\ref{fig:challenge}-D) and multiple-view visual analytics systems (\figurename\ref{fig:challenge}-C).
% The annotation of the large quantity of images needs massive efforts. 
Third, it is also challenging to ensure the quality of the annotations, given the diverse knowledge expertise of annotators and the lack of ``the ground truth''. Addressing conflicts and reducing biases is critical for data annotation.

To address the first challenge, we use a comprehensive taxonomy proposed by Borkin et al.~\cite{borkin2013makes} and annotate the visualizations by regarding them as compositions of different visualizations~\cite{javed2012composite}. 
% The taxonomy covers most of the visualizations that appeared in visualization publications.
Based on the taxonomy, we invite visualization practitioners to annotate the visualization types.
For the second challenge, we set up a series of criteria for decomposing and localizing visualizations in the images. 
Based on the criteria, we recruit trained crowd workers to annotate the bounding box for each visualization. 
To tackle the third challenge, we adopt a series of measures for quality control, including the gold standard~\cite{su2012crowdsourcing}, majority voting, and sampling test.
Our contributions are threefold. 
\begin{itemize}
    \item We build a novel dataset named VisImages from IEEE VAST and InfoVis for the research of literature analysis and AI4VIS.
    % Each image is accompanied by a textual caption and a series of bounding boxes representing the positions and types of the visualizations contained. 
    We release the dataset and codes for data collection at \textcolor{blue}{\url{https://visimages.github.io/}}. 
    \item We present an overview of the use of visualizations in visualization publications and compare it with the images collected from public sources. From the analysis, we gain insights into the peculiarities of visualizations in academic publications.
    \item We showcase the usefulness of VisImages through three use cases, namely, 1) exploring and analyzing the evolution of visualizations in publications with VisImages Explorer, 2) evaluating the generalizability and robustness of visualization classification models, and 3) localizing visualizations in the interfaces of visual analytics systems automatically.
\end{itemize}
\section{Related Work}
This section introduces related studies on visualization datasets and visualization literature analysis.

\subsection{Image Datasets in Visualization}
We first introduce existing visualization image datasets and demonstrate that VisImages can greatly boost the research in visualization because of its unique data source, visualization publications. In addition, VisImages is, to our best knowledge, the most comprehensive one regarding visualization quantity, layout complexity, label types, etc.

The visualization community has built a variety of image datasets of basic charts (such as bar charts or scatterplots) with simple layouts.
The images in these datasets are mainly collected from the Internet, such as social media (e.g., Twitter) and media outlet (e.g., BBC), or generated by visualization libraries (e.g., D3~\cite{2011-d3}, Vega-Lite~\cite{satyanarayan2016vega}).
For example, Battle et al.~\cite{battle2018beagle} gathered over 41,000 SVG-based charts, manually labeled each chart one of 24 chart types, and trained classification models to analyze the chart distribution on the web.
Jung et al.~\cite{jung2017chartsense} collected 5,659 images consisting of 10 chart types to develop models for chart classification and proposed ChartSense for data extraction.
Savva et al.~\cite{savva2011revision} delivered a dataset containing 2,601 single-chart images in 10 categories.
The dataset is used to develop a system called ReVision to redesign the charts for better visual styles.
Similarly, Poco et al.~\cite{poco2017reverse} collected more than 5,000 bitmap images and annotated chart types (area, line, bar, and scatter) and textual annotations (labels and titles of axes and legend) of the charts.
The dataset is used for reconstructing the original charts with declarative grammars, such as Vega-Lite~\cite{satyanarayan2016vega}.
Borkin et al.~\cite{borkin2013makes, borkin2015beyond} developed MassVis for memorability study. 
They collected more than 2,000 single-chart visualizations and categorized them into 12 categories.
For each image, they evaluated the data-ink ratio and visual density through crowdsourcing.
They also annotated a subset of 396 images for detailed information (e.g., annotations, axis, and data).
Lee et al.~\cite{lee2017viziometrics} collected the images from scientific publications and categorized them into equation, photo, diagram, etc.
% Aforementioned datasets mainly focus on the images of relatively simple visualizations, such as single-view visualizations created by visualization libraries. 

\begin{table*}[!htb]
    \centering
    \caption{Existing visualization datasets in the visualization community.}
    \label{tab:datasets}
    \scriptsize
    \begin{tabularx}{\linewidth}{cccccccccX}
    % \toprule
    % \textbf{Dataset} & VisImages & Beagle & MassVis & REV\\\midrule
    % \textbf{Source} & IEEE VAST \& InfoVis & D3, Chartblocks, Fusion Charts, Graphiq, Plotly & science, infographics, news, and government& Vega, Quartz, and academic paper\\\midrule
    % \textbf{Layout} & multi-chart & single-chart & single-chart & single-chart\\ \midrule
    % \textbf{\#Vis./\#Img.} & 35,096/12,267 & 33,778/33,778 & 2000+ (396)/2000+ & 5000+/5000+\\\midrule
    % \textbf{\#Cat.} & 12 (\underline{30}) & 24 & \underline{12} (63) & 4\\ \midrule
    % \textbf{Content} & category, position, and caption & category & 2000+: category, data-ink, density; 396: legend, title &  visual marks and textual elements\\ \bottomrule

    \toprule
    \multirow{2}{*}{\textbf{Dataset}} & \multirow{2}{*}{\textbf{Audience}} & \multirow{2}{*}{\textbf{Layout}} & \multirow{2}{*}{\textbf{\makecell[c]{\#Annotated\\Visualizations}}} & \multirow{2}{*}{\textbf{\#Images}} & \multirow{2}{*}{\textbf{\makecell[c]{\#Annotated\\Categories}}} & \multicolumn{3}{c}{\textbf{Label Types}} &\multirow{2}{*}{\textbf{How to label?}}\\
    &&&&&& type    & bbox   & caption &   \\\midrule
    MassVis~\cite{borkin2013makes} & general users & single & $\sim$2000 & $\sim$2000& 12 & $\checkmark$& -&- &manual annotation\\\midrule
    REV~\cite{poco2017reverse} & general users & single & $\sim$5000 & $\sim$5000 & 4& $\checkmark$ &- & -& machine generation + manual refinement\\\midrule
    Beagle~\cite{battle2018beagle} & general users & single & 33,778 & 33,778 & 24 & $\checkmark$& -& -&manual annotation\\\midrule
    ChartSense~\cite{jung2017chartsense} & general users & single & $\sim$2000 &$\sim$2000 & 10 & $\checkmark$& -&- & search engine + manual refinement\\ \midrule
    VizioMetrics~\cite{lee2017viziometrics} & general users & - & - &\textbf{$\sim$4,986,302*} & 5 & $\checkmark$&-& - &object classfication + manual annotation\\\hline\midrule
    VIS30K~\cite{chen2021vis30k} & visualization experts & - & - &$\sim$30,000 & 4 & $\checkmark$&-& - &object detection  + manual refinement\\\midrule
    MV Dataset~\cite{chen2020composition} & visualization experts & multiple & not reported &360 & 14 & $\checkmark$&$\checkmark$& - &manual annotation\\\hline\midrule
    \textbf{VisImages} & \textbf{visualization experts*} & \textbf{multiple*} & \textbf{35,096*}&12,267 & \textbf{34*} & $\checkmark$&$\checkmark$&$\checkmark$&manual annotation\\\bottomrule
    \end{tabularx}
\end{table*}

Other than basic charts, the images in visualization publications contain novel designs created by visualization experts, such as system interfaces, which are the interests many studies lay in.
For example, Li et al.~\cite{li2018toward} collected images from IEEE SciVis and conducted user studies to understand the relation between memorability and image characteristics.
Chen et al.~\cite{chen2021vis30k} adapted object detection models (Faster-RCNN~\cite{ren2015faster} and YOLOv3~\cite{redmon2018yolov3}) to crop the figures and tables from visualization publications and proposed VIS30K, a dataset of figures and tables.
Similarly, Zeng et al.~\cite{zeng2021vistory} collected figures from IEEE VIS to visualize and analyze the evolution of figures.
However, detailed information of the visual designs, such as chart types, chart positions, and captions, are not considered in their dataset.
% \dazhen{highlight our differences and novelty over VIS30K, but not criticize the VIS30K.}
Chen et al.~\cite{chen2020composition} collected figures of multiple-view visualizations (MVs) from the publications and annotated the view positions and types.
They contributed a corpus of 360 MV images and statistically analyzed the view types and layouts.
We conduct a detailed comparison between VisImages and the existing datasets from the perspectives of target audience, visualization layout, quantity, category, and label type, as illustrated in \tablename~\ref{tab:datasets}.

% VisImages focuses on the images from visualization publications, but is different from aforementioned datasets in two aspects.
% First, VisImages consists of images with varying types and layouts, such as multiple-view visualizations (e.g., system interfaces), single-view visualizations with novel designs (e.g., visualizations in \figurename\ref{fig:challenge} (C, D)), and simple charts (e.g., experiment results).
% On the contrary, Li et al.~\cite{li2018toward} mainly focused on volume renderings in SciVis and Chen et al.~\cite{chen2020composition} focused on system interface.
% Second, VisImages exceeds existing datasets in the granularity of annotations.
% For example, compared to VIS30K, we not only extracted the figures but also annotated the types and positions of the charts contained.
% Therefore, VisImages is a good complement to these datasets in supporting diverse machine learning models to visualization scenarios, such as visualization localization and classification.
% \subsection{VisImages vs. Other Visualization Datasets.}

VisImages is novel for visualization research compared to existing datasets because of its usefulness for visualization understanding and its usability due to comprehensive annotations.
First, VisImages comprises fresh designs with complex configurations that are created by visualization researchers. Meanwhile, existing datasets, such as VizioMetrics~\cite{lee2017viziometrics}, Beagle~\cite{battle2018beagle}, ChartSense~\cite{jung2017chartsense}, MassVis~\cite{borkin2015beyond, borkin2013makes}, and REV~\cite{poco2017reverse}, collect data from public sources and contain a majority of basic charts.
The unique data source makes VisImages a compelling dataset for studying the variations or combinations of common visualization types as well as innovative visual designs.
% Moreover, our decomposition and indexing of the visualizations in the visual designs facilitate quick searching and exploration of frontier designs for visualization researchers.
Besides, the novel visual designs in VisImages serve as a challenging benchmark for new research methods that boost the study of AI4VIS~\cite{wu2021survey}.

Second, VisImages is outstanding in its comprehensive annotations, which are well-suited for various scenarios.
1) \textbf{Annotation Granularity}: Compared to VIS30K~\cite{chen2021vis30k} and VizioMetrics~\cite{lee2017viziometrics} that classify the images by their goals or usages (e.g., diagram, equation, photo, plot, and table) at the image level, VisImages further specify the visualization types and bounding boxes within the images with a fine-grained taxonomy.
2) \textbf{Design Complexity:} Compared to MassVis~\cite{borkin2015beyond, borkin2013makes} and REV~\cite{poco2017reverse} that are mainly composed of single-chart visualizations, VisImages includes detailed annotations on multi-chart visualizations, such as visual analytics systems.
\noindent3) \textbf{Visualization Quantity:} In addition to the complexity, VisImages also covers a large quantity of visualizations (about 100x of the quantity of MV Dataset~\cite{chen2020composition}), which can fulfill the requirements of training deep learning models.
With the above features, VisImages can be directly used in many scenarios, such as classification and localization, which is practical for the research of AI4VIS.
% Therefore, VisImages is a good complementary to existing datasets, providing comprehensive annotations for the images from visualization publications.

\subsection{Visualization Literature Analysis \& Datasets}
Literature analysis is an important research area for indexing and understanding the publications.
Current studies mainly use the following four types of data: text, citations, authors, and metadata~\cite{ federico2016survey}.

Many datasets of visualization publications \cite{fekete2004infovis, xie2016visualizing, plaisant2007promoting, cook2014vast, isenberg2016vispubdata} are used to support interactive literature analysis.
The most up-to-date one is vispubdata.org~\cite{isenberg2016vispubdata}, which contains metadata of publications in IEEE VIS sub-conferences. 
The publication data, such as authors, references, and keywords, is collected from the electronic proceedings.
A series of visual analytics tools, such as CiteVis2, CiteMatrix, and VisList~\cite{isenberg2016vispubdata}, were proposed on the basis of vispubdata.org.
Ponsard et al.~\cite{ponsard2016paperquest} proposed PaperQuest, which is a tool to search for relevant papers that users are interested in.
Several studies~\cite{chuang2013topic, isenberg2016keyvis} also attempt to organize publications based on research topics.
Chuang et al.~\cite{chuang2013topic} introduced a framework to use topic modeling for the analysis of InfoVis corpus.
Isenberg et al. proposed KeyVis~\cite{isenberg2016keyvis} that extracts the keywords of visualization papers. 
However, none of the above studies investigate the image data.
VisImages comprises a large dataset of images with rich annotations, which can provide additional dimensions for literature analysis.

\section{Dataset Construction}
In this section, we overview the construction of VisImages.

\label{sec:tax}
\begin{figure*}[tb]
    \centering
    \includegraphics[width=\linewidth]{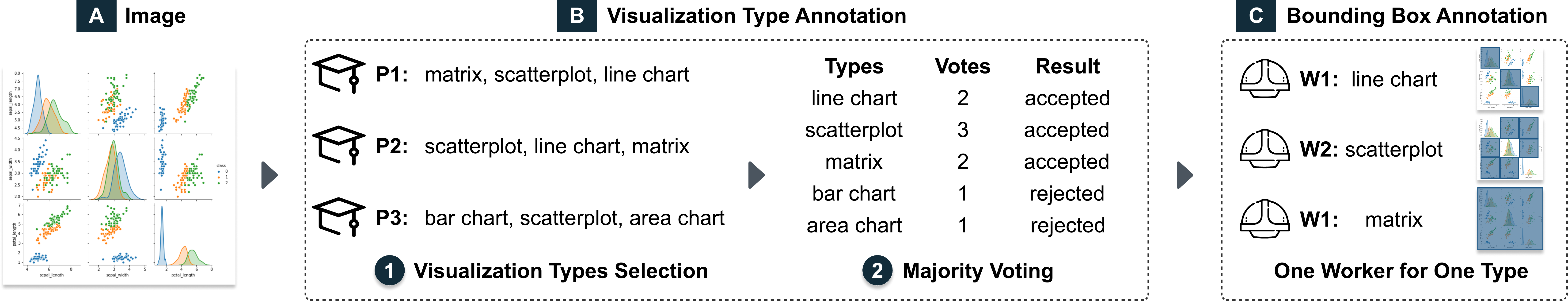}
    \caption{The pipeline of data annotation. (A) shows an image sample for annotation. (B) shows the process of annotating visualization types, in which three visualization practitioners independently specify the visualization types that the image contains. (C) shows the process of majority voting, in which we accept the visualization types that receive at least two votes from three practitioners. (D) shows the process of annotating bounding boxes, in which each crowd worker focuses on one type of visualization and draws the bounding boxes to specify the positions of visualizations.}
    \label{fig:pipeline}
\end{figure*}

\begin{table}[tb]
    \centering
    \scriptsize%
    \caption{Visualization Taxonomy.}
    \begin{tabularx}{\linewidth}{lX}
    \toprule
    \textbf{Categories} & \textbf{Sub-types} \\\midrule
    Area& area chart, proportional area chart (PAC)\\ \midrule
    Bar & bar chart \\ \midrule
    Circle  & donut chart, pie chart\\ \midrule
    Diagram & flow diagram, chord diagram, Sankey diagram, Venn diagram\\ \midrule
    Statistic& box plot, error bar, stripe graph\\ \midrule
    Table & table\\ \midrule
    Line& contour graph, line chart, storyline, polar plot, parallel coordinate (PCP), surface graph, vector graph\\ \midrule
    Map & map\\ \midrule
    Point& scatter plot\\ \midrule
    Grid \& Matrix & heatmap, matrix\\ \midrule
    Text& phrase net, word cloud, word tree\\ \midrule
    Graph \& Tree  & graph, tree, treemap, hierarchical edge bundling (HEB), sunburst/icicle plot \\ \midrule
    Special  &  glyph-based visualization, unit visualization\\ \bottomrule
    \end{tabularx}
    \label{tab:tax}
\end{table}

\subsection{Data Preprocessing}
To construct VisImages, we started with collecting images from top-venue visualization publications (\figurename\ref{fig:pipeline}-A).
In this study, we focused on 2D static visualizations and collected the images from VAST (IEEE Conference on Visual Analytics Science \& Technology) and InfoVis (IEEE Conference on Information Visualization). 
We excluded SciVis (IEEE Conference on Scientific Visualization) papers since these papers generally comprise a large number of images depicting the results of 3D rendering, which are beyond the scope of this paper. 
We firstly downloaded PDF files of the papers according to the digital object identifier (DOI) provided in vispubdata.org~\cite{isenberg2016vispubdata}. 
Next, we used PDFFigures 2.0~\cite{clark2016pdffigures} to extract images and captions from the files, and manually checked and corrected the results. 
We focused on the figures and tables indexed by \textit{Figure} and \textit{Table} and inline figures without a caption.
In total, we processed 1,397 papers in VAST and InfoVis dated from 1996 to 2018 and collected 12,267 images with 12,057 textual captions.

\subsection{Visualization Taxonomy}
To classify visualizations, we used the taxonomy proposed by Borkin et al.~\cite{borkin2013makes} that categorizes visualizations into a two-level structure, i.e., 12 categories with sub-types.
The taxonomy classifies the visualizations used in the public (i.e., infographics, news media, scientific journals, and government \& world organization) based on their visual encodings (e.g., bar and area), visual tasks (e.g., statistics), and visual layouts (e.g., diagram).
However, we discovered that the original taxonomy has some sub-types that are similar in definition, such as bar chart and histogram.
To avoid ambiguities, we merge these types together.
In addition, we also discovered that some visualization types are not listed in the taxonomy, such as icicle plot and glyph-based visualizations.
Therefore, we added these types into the original taxonomy.
The taxonomy used in our dataset consists of 13 categories and 34 sub-types, as shown in \tablename{\ref{tab:tax}}.

\subsection{Pipeline of Data Annotation}
To annotate the visualization types and bounding boxes in the images, we designed a pipeline that takes advantage of the expertise of visualization practitioners and the scalability of crowdsourcing.

We first recruited qualified participants from the university for visualization type annotation given the visualization expertise required for recognizing the types of visualizations (\figurename~\ref{fig:pipeline}-B)).
To identify different visualizations, we basically used a visualization taxonomy proposed by Borkin et al.~\cite{borkin2013makes}. 
Specifically, for each image, three participants were asked to identify all possible visualization types that appear (\figurename\ref{fig:pipeline}-B1).
To address conflicts, we adopted majority voting that a visualization type was accepted only if at least two participants voted for it (\figurename\ref{fig:pipeline}-B2).
After this step, we collected 10,289 images containing visualization types within our taxonomy.
The details of visualization type annotation are illustrated in Section~\ref{sec:type-annotation}.

Second, we annotated the bounding boxes of visualizations in the images through crowdsourcing (\figurename\ref{fig:pipeline}-C). 
To ensure high data quality, we carefully designed the tasks and performed cross-validation (Section~\ref{sec:bbox-annotation}). 
As a result, we obtained a dataset of 35,096 bounding boxes, each corresponding to a specific visualization.
The detailed procedure of bounding box annotation is demonstrated in Section~\ref{sec:bbox-annotation}.

\section{Process of Data Annotation}
\label{sec:annotation}
We label the visualization types and bounding boxes based on the refined taxonomy and annotation pipeline.

\subsection{Visualization Type Annotation}
\label{sec:type-annotation}
Identifying visualizations and their variations is challenging and requires extensive knowledge of visualization. 
Thus, we recruited the researchers and students who were experienced in visualization research to annotate the visualization types that appear in the images.
Please note that the term ``type'' refers to the visualization sub-types in \tablename{\ref{tab:tax}}.

\textbf{Participants.}
We recruited 25 participants, including 1 senior visualization expert who had six-year experience in visualization research, 
13 Ph.D. candidates with the research interest of visualization, 
7 master students majoring in information visualization, 
and 4 undergraduate students who had taken the undergraduate course of data visualization.
Most of them (15/25) had published papers in IEEE VIS.

\textbf{Procedure.}
The annotation procedure consisted of a training session and a formal study.
In the training session, we first introduced the taxonomy and the definition of each visualization sub-type with examples.
Then we introduced the details of annotation tasks. 
Specifically, annotating visualization types for an image is a multi-label task in our study.
In such a task, a participant was shown an image and asked to select all the visualization sub-types occurring in the image based on our taxonomy.
If the participant thought the image does not contain the visualization types within our taxonomy, they could choose the additional option ``others.''
After the introduction, participants were asked to take a test to ensure that they had correctly understood the taxonomy. 
The test contained 20 images covering all visualization types (an image might include multiple types), and participants were considered eligible for the formal study only if they correctly annotated more than 18 images. 
All participants passed the test at their first attempt.

In the formal study, all participants annotated the images independently.
Before the annotation, the participants had to enter their names as identification. 
All annotated data recorded the name of the participant. 
During the annotation, the participants could not see the results of others. 
We developed and deployed an online interface for data annotation. 
The data was stored in a backend server, which recorded the annotation logs and managed the task assignments.

Each round of annotation comprised 40 tasks. 
It took about 10 minutes to complete a round.  
Each participant was assigned at most 40 rounds.
To avoid overloading, participants were allowed to accomplish all images within five days.
We paid \$0.05 for each accepted task.

\textbf{Quality Control}.
We adopted the methods of gold standards and majority voting for quality control.
The gold standards were the images manually selected and inserted into each round to test whether the participants were focusing on the tasks.
The gold standard images contained simple charts placed in an obvious position, and the participants were expected to correctly specify the visualization types easily.
Each round included eight images with gold standards.
If a participant failed in more than one gold standard in a round, all results from this round would be rejected, and we would reassign these images to other participants.
Finally, we obtained 940 accepted rounds and 102 rejected rounds.
In the accepted rounds, the participants obtained an accuracy of 96.4\% on the gold standards.

In addition, we used majority voting to address ambiguities in the annotation.
Specifically, each image would be annotated by three participants independently.
The selection of a visualization sub-type by a participant would be regarded as a ``vote''. 
For each image, the sub-types with at least two votes would be accepted as the fact that the image contained the instances of these visualization sub-types.
Otherwise, the sub-types would be suspended for further discussion.
Due to the majority voting, the entire annotation process contained at least 12,267 images $\times$ 3 repetitions = 36,801 annotations.
We computed the acceptance rate of a participant to evaluate the similarity between his/her annotations and the accepted results. Overall, the participants gained an average acceptance rate of 88.3\%.
In addition, we compute the intercoder reliability with Krippendorff's Alpha-Reliability~\cite{krippendorff2011computing}, which supports multiple labels and multiple observers in an annotation task.
We compute the alpha value using the Python implementation in ``\textit{nltk.metrics.agreement}''~\cite{agreementmetric}. Finally, we obtain the intercoder reliability value of 76.8\%.

Finally, we found 10,289 out of 12,267 images were assigned labels of visualization sub-types.
We investigated the rest of images and discovered that these images are usually the ones with low resolution, the ones explaining methods, models, or algorithms, the ones of photos, 3D rendering, or artistic work, and the ones with pure text. 
Therefore, we assigned the rest with a label of ``others.''
The distribution of each sub-type is shown in \tablename~\ref{tab:distribution}.

\subsection{Bounding Box Annotation}
\label{sec:bbox-annotation}
With the specified visualization sub-types in each image, we further annotated bounding boxes (i.e., the positions in the images) for these visualizations.
To improve efficiency, we employed crowd workers from a professional data annotation company, who are well-trained for drawing bounding boxes for machine learning tasks. 

\begin{figure*}[tb]
    \centering
    \includegraphics[width=1\linewidth]{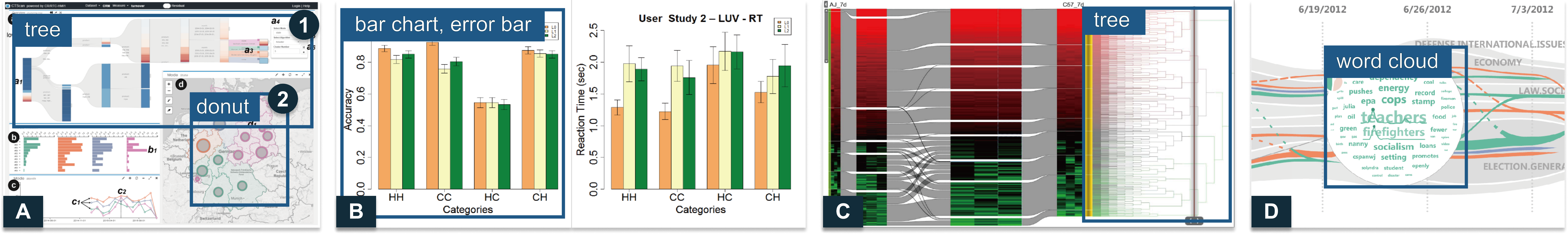}
    \caption{Criteria for bounding box drawing. (A) shows how to draw bounding boxes for an independent visualization (A1) and multiple identical visualization sub-types (A2)~\cite{liu2018tpflow}. (B) shows the bounding boxes of multiple sub-types (bar charts and error bars) which are included in the same coordinate. (C) shows how to draw bounding boxes for visualizations (tree) which are closely connected to other visualizations~\cite{lex2010comparative}. (D) shows the bounding box of the target visualization (word clouds) which is overlaid on another visualization~\cite{xu2013visual}}
    \label{fig:criteria}
\end{figure*}

\textbf{Criteria.}
Our criteria are based on the composition of the visualizations, i.e., visualization with coordinates or without coordinates.
For a visualization with coordinates, the bounding box should cover all components of the coordinates, e.g., axis name, axis labels, chart title, and legends, if they are close to the main bodies of visual representations (\figurename\ref{fig:criteria}-B). 
If multiple sub-types share the same coordinate (e.g., error bar \& bar chart in \figurename\ref{fig:criteria}-B), the area of their bounding boxes should be the same.
For the visualizations without coordinates, we identify two situations, i.e., 1) independent visualizations without any connection or overlapping with other visualizations and 2) the visualizations connected to or overlapped with other visualizations.
For the first case, the contents are the visualization itself (\figurename\ref{fig:criteria}-A1).
For the second case, we only focus on the contents of the requested sub-type. 
For example, the tree in \figurename\ref{fig:criteria}-C is connected to the Sankey diagram, and the word cloud in \figurename\ref{fig:criteria}-D overlays on the area charts.
The bounding boxes should only cover the contents of the tree and word cloud, respectively.
In addition, there is an exception that requires further specification.
Some visualizations contain multiple smaller visualizations of identical sub-type (e.g., the donut charts in the map in \figurename\ref{fig:criteria}-A2).
In this case, when the number is larger than 10, we annotate them integrally with a single box with a label of ``small multiples.''

\textbf{Procedure.}
The annotation procedure consisted of a training session and a formal annotation. 
To reduce the training load, each crowd worker was asked to focus on only one sub-type.
Therefore, in the training session, a crowd worker was introduced one specific visualization sub-type, including the definition, examples, and the above annotation criteria. 
After that, the crowd workers were asked to take a test to ensure that they understood the sub-type and requirements. 
Only crowd workers who passed the test proceeded to the formal annotation. 
They were then assigned images containing specific visualization sub-types and required to draw the bounding boxes for this type of visualization.
During the annotation, sampling tests were adopted to ensure quality.

\begin{figure}
    \centering
    \includegraphics[width=\linewidth]{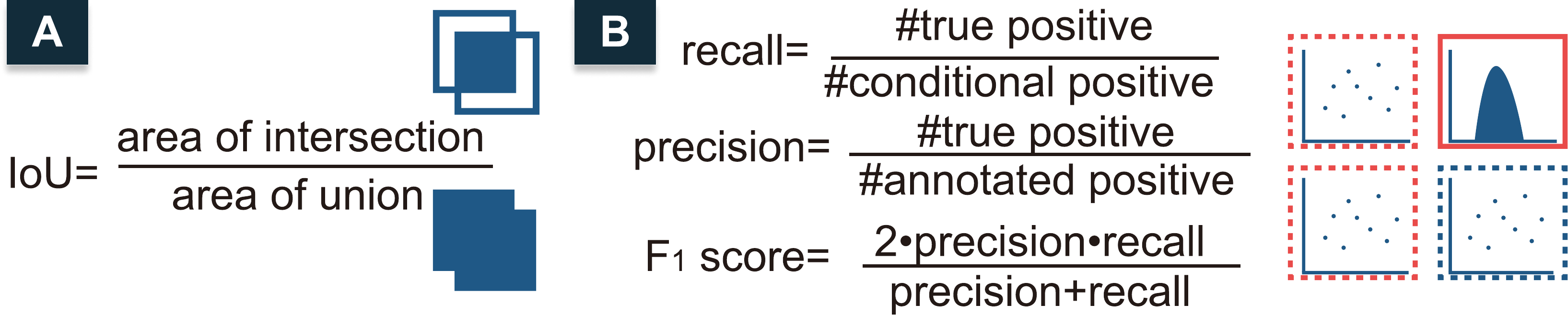}
    \caption{Quality measurement for bounding box annotation. (A) shows how to compute the IoU of two bounding boxes. (B) shows the computation of recall, precision, and $F_1$ score. Red boxes represent the boxes labeled by crowd workers (annotated positive), dotted boxes represent the ground truth (condition positive), and red-dotted boxes are the ground truth correctly labeled by the crowd workers (true positive).}
    % \vspace{-10pt}
    \label{fig:metric}
\end{figure}
\textbf{Quality Measurement \& Control.}
We evaluated the correctness of bounding boxes and tasks to control the annotation quality.
The correctness of a bounding box was measured by intersection over union~\cite{Redmon2016YouOL} (IoU, illustrated in \figurename\ref{fig:metric}-A) with the ground-truth bounding box.
Only when the IoU of the bounding box and the ground truth is higher than 0.9, the bounding box was accepted.
Besides, the quality of a series of tasks was measured by the $F_1$ score, a metric balancing the recall and precision.
The calculation of recall, precision, and $F_1$ score is presented in \figurename\ref{fig:metric}-B.
\label{sec:iou}

To ensure quality, we adopted a sampling test on both batch level and worker level.
We divided the 10,289 images equally into five batches and performed annotations batch by batch.
The batch-level sampling test was performed after completing a batch of annotations.
We randomly sampled 10\% of the results and evaluated the $F_1$ score.
If the $F_1$ was lower than 95\%, the whole batch of annotation would be rejected.
The rejected batch would be annotated again until the $F_1$ score reached 95\%.
The worker-level sampling test was conducted during one batch of annotations, where 15\% annotations of a worker would be randomly sampled for $F_1$ score evaluation.
If the $F_1$ was lower than 95\%, all finished tasks of this worker in this batch would be rejected and annotated again.
For the workers who failed the sampling test, their sampling rate would increase by 5\% at the next test.
Each accepted bounding box was paid with 0.03\$.
\section{VisImages}
In this section, we present an overview of VisImages data and compare the distribution of visualizations in VisImages with that from other sources~\cite{borkin2013makes}. 

\begin{table}[tb]
    \centering
    \scriptsize%
    \caption{Distribution of the visualization sub-types.}
    \begin{tabular}{llllll}
        \toprule
        % bar chart &\textChart{5000} & \textChart{2000}\\
        \textbf{Sub-type} & \textbf{\#bbox}&\textbf{\#img}& \textbf{Sub-type} & \textbf{\#bbox}&\textbf{\#img.}\\\midrule
        bar chart&\textChart{5053}&\textChart{2058}&pie chart&\textChart{371}&\textChart{153}\\
        scatterplot&\textChart{4269}&\textChart{1754}&PAC&\textChart{288}&\textChart{130}\\
        graph&\textChart{3722}&\textChart{1615}&box plot&\textChart{277}&\textChart{147}\\
        heatmap&\textChart{3202}&\textChart{1187}&unit visualization&\textChart{275}&\textChart{107}\\
        line chart&\textChart{3004}&\textChart{1300}&sunburst/icicle&\textChart{260}&\textChart{120}\\
        table&\textChart{2172}&\textChart{1676}&sankey diagram&\textChart{260}&\textChart{147}\\
        map&\textChart{2106}&\textChart{986}&stripe graph&\textChart{239}&\textChart{123}\\
        matrix&\textChart{1611}&\textChart{656}&HEB&\textChart{185}&\textChart{61}\\
        tree&\textChart{1292}&\textChart{667}&chord diagram&\textChart{128}&\textChart{72}\\
        area chart&\textChart{1125}&\textChart{527}&polar plot&\textChart{123}&\textChart{56}\\
        flow diagram&\textChart{1118}&\textChart{873}&storyline&\textChart{46}&\textChart{25}\\
        PCP&\textChart{975}&\textChart{541}&contour graph&\textChart{16}&\textChart{12}\\
        error bar&\textChart{709}&\textChart{342}&surface graph&\textChart{13}&\textChart{7}\\
        treemap&\textChart{554}&\textChart{268}&word tree&\textChart{9}&\textChart{9}\\
        glyph-based&\textChart{523}&\textChart{259}&phrase net&\textChart{7}&\textChart{7}\\
        word cloud&\textChart{392}&\textChart{184}&Venn Diagram&\textChart{4}&\textChart{4}\\
        donut chart&\textChart{376}&\textChart{143}&vector graph&\textChart{4}&\textChart{2}\\\midrule
        others&-&\textChart{1978}\\
        \bottomrule
    \end{tabular}
    \label{tab:distribution}
\end{table}

\begin{figure*}[tb]
    \centering
    \includegraphics[width=\textwidth]{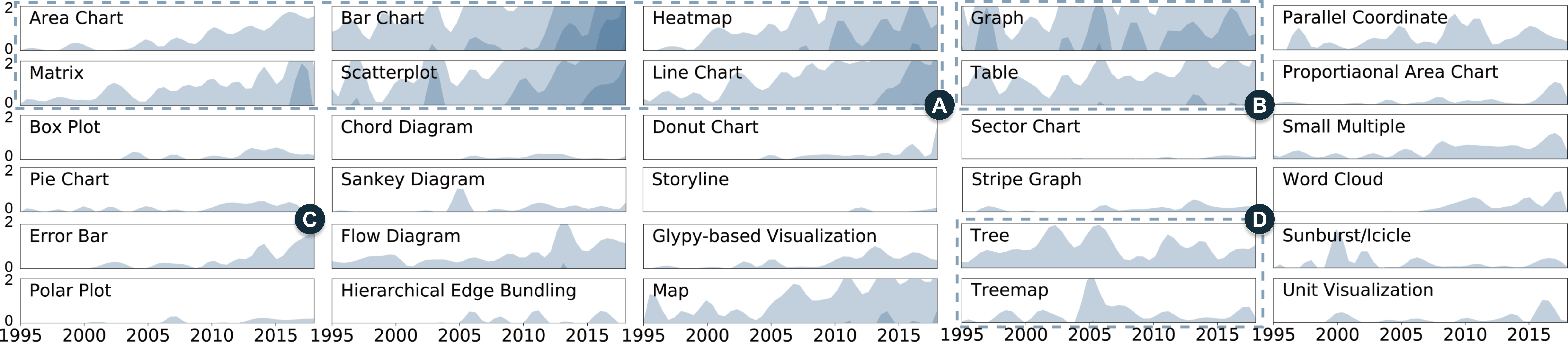}
    \caption{Horizon charts demonstrate the average numbers of visualization bounding boxes in a paper over years.}
    \label{fig:horizon}
\end{figure*}
\subsection{Overview of the Data}
VisImages contains 12,267 images from 22-year VAST and InfoVis publications with 12,057 captions and 35,096 visualization bounding boxes.
\tablename~\ref{tab:distribution} shows the numbers of images (\#img) and bounding boxes (\#bbox) of each sub-type.
For the frequent types, we observe that the number of bounding boxes of some sub-types (e.g., bar chart and scatterplot) is about two times more than the number of images. 
That is, multiple instances of these sub-types appear in one image simultaneously. 
For bar charts and scatterplots, the reason might be that they are basic charts and commonly serve as units of small multiples (e.g., scatterplot matrix).
On the contrary, tables and flow diagrams have similar numbers of bounding boxes and images. 
We find that they usually occupy the entire image, since the tables are used independently to show the results of experiments or studies, and the flow diagrams are used to show the pipeline or framework of the methods.

\figurename{\ref{fig:horizon}} depicts the distribution of each sub-type from 1996 to 2018 using horizon charts, with color darkness encoding the number of bounding boxes.
Several visualization sub-types are becoming increasingly popular, such as bar chart, area chart, scatterplot, matrix, line chart and heatmap (\figurename{\ref{fig:horizon}}-A).
We notice that the dark area of graph visualizations distributes evenly across years (\figurename{\ref{fig:horizon}}-B), indicating that graph visualization has long been frequently used in the visualization community~\cite{han2021netv, wang2021g6}. 
Similarly, tables have always been a common visualization type in publications (\figurename{\ref{fig:horizon}}-B).
Besides, we observe that the area of treemap becomes abruptly larger in 2005 while the area of tree reaches a peak in 2003 and 2005 (\figurename{\ref{fig:horizon}}-D). 
The increase of tree and treemap visualizations implies a more popular investigation in hierarchical data~\cite{schulz2011treevis}.
Moreover, the number of error bars continuously increases in recent years (\figurename{\ref{fig:horizon}}-C), indicating that statistical analysis on the error is increasing, such as user studies and model experiments.

\subsection{Comparing Visualizations in Different Fields}
Furthermore, we analyze the visualization distribution in academic visualization publications and materials that target to general audience.
We use the statistics in MassVis~\cite{borkin2013makes} for comparison (\tablename~\ref{tab:comparison}), which collects images from four different sources, i.e., scientific publications (Nature), infographics, news, and government \& world organization.

The comparison can be conducted directly because the taxonomies are the same at the category level, as shown in \tablename~\ref{tab:tax}.
From \tablename~\ref{tab:comparison}, we notice that the distribution in visualization publications is more balanced compared to the others. 
\textit{Graph and Tree} occupy the largest share in visualization publications, which do not frequently appear in other sources.
The reason might be that a quantity of research in our community focuses on visualizing hierarchical or network data.
On the other hand, news media and government \& world organizations prefer basic visual representations such as \textit{Bar}, \textit{Table}, and \textit{Line} because the data they mostly present is relatively simple and in the form of tabular.
Scientific papers prefer \textit{Diagram}, \textit{Line}, and \textit{Point} for the presentation of methodology and experiment results.
We notice that \textit{Text}, which includes word clouds, word trees, and phrase nets, accounts for a portion in visualization publications but rarely appears in other sources.
A lot of visualization research investigates variations of word cloud to make it more informative and effective, such as ManiWordle~\cite{koh2010maniwordle} and dynamic word cloud~\cite{cui2010context}.
However, in public, given that the most commonly used media is text, the authors might expect to use graphical elements other than text to improve the expressiveness.

\begin{table}[tb]
    \centering
    \caption{The distribution of visualizations in VisImages and MassVis.}
    \scriptsize%
    \begin{tabular}{llllll}
        \toprule
        \multirow{2}{*}{Source} & \multirow{2}{*}{VisImages} & \multicolumn{4}{c}{MassVis} \\\cmidrule{3-6}
          & & Scientific& Infographics& News& Government\\\midrule
        Area& \Chart{4.0}&\Chart{1.9}&\Chart{4.4}&\Chart{4.4}&\Chart{3.5}\\
        Bar& \Chart{12.1}&\Chart{6.4}&\Chart{5.9}&\Chart{40.2}&\Chart{36.9}\\
        Circle& \Chart{1.8}&\Chart{0.3}&\Chart{4.7}&\Chart{1.3}&\Chart{6.6}\\
        Diag.& \Chart{6.3}&\Chart{27.4}&\Chart{30.6}&\Chart{7.2}&\Chart{5.0}\\
        Stat.& \Chart{3.7}&\Chart{3.2}&\Chart{0.3}&\Chart{0.3}&\Chart{1.3}\\
        Table& \Chart{9.8}&\Chart{8.3}&\Chart{32.8}&\Chart{8.2}&\Chart{21.5}\\
        Line& \Chart{11.2}&\Chart{19.1}&\Chart{1.6}&\Chart{19.1}&\Chart{12.9}\\
        Map& \Chart{6.0}&\Chart{9.2}&\Chart{9.1}&\Chart{13.5}&\Chart{7.3}\\
        Point& \Chart{10.6}&\Chart{16.6}&\Chart{2.8}&\Chart{5.0}&\Chart{0.5}\\
        Grid& \Chart{7.2}&\Chart{2.5}&\Chart{1.9}&\Chart{0}&\Chart{0}\\
        Text  & \Chart{1.1}&\Chart{0}&\Chart{0}&\Chart{0.5}&\Chart{0}\\
        Graph& \Chart{16.7}&\Chart{5.1}&\Chart{5.9}&\Chart{0.3}&\Chart{0}\\
        Special  & \Chart{9.5}&&&&\\\midrule
        \#Vis.& 10,289&348&490&704&528\\
        \bottomrule
    \end{tabular}
    \label{tab:comparison}
\end{table}
\section{Use Cases}
\label{sec:cases}
We present three use cases in this section to show the usefulness of VisImages. 
Specifically, the first case demonstrates the use of all metadata and annotations, the second case demonstrates the use of visualization type data, and the third case demonstrates the use of bounding box data.
\begin{figure*}[!tb]
    \centering
    \includegraphics[width=\linewidth]{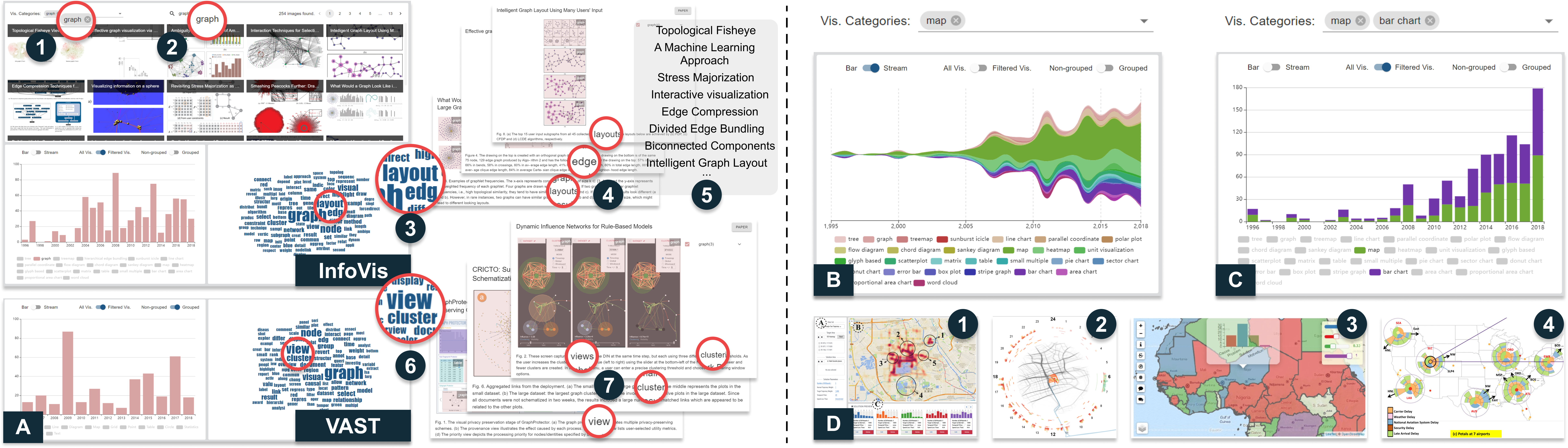}
    \caption{Usage scenarios for junior visualization researchers and novice visualization designers: reviewing papers of graph visualizations (A) and inspiring visual designs (B, C, D).}
    \label{fig:interface-case}
\end{figure*}

\begin{figure}[!tb]
    \centering
    \includegraphics[width=\linewidth]{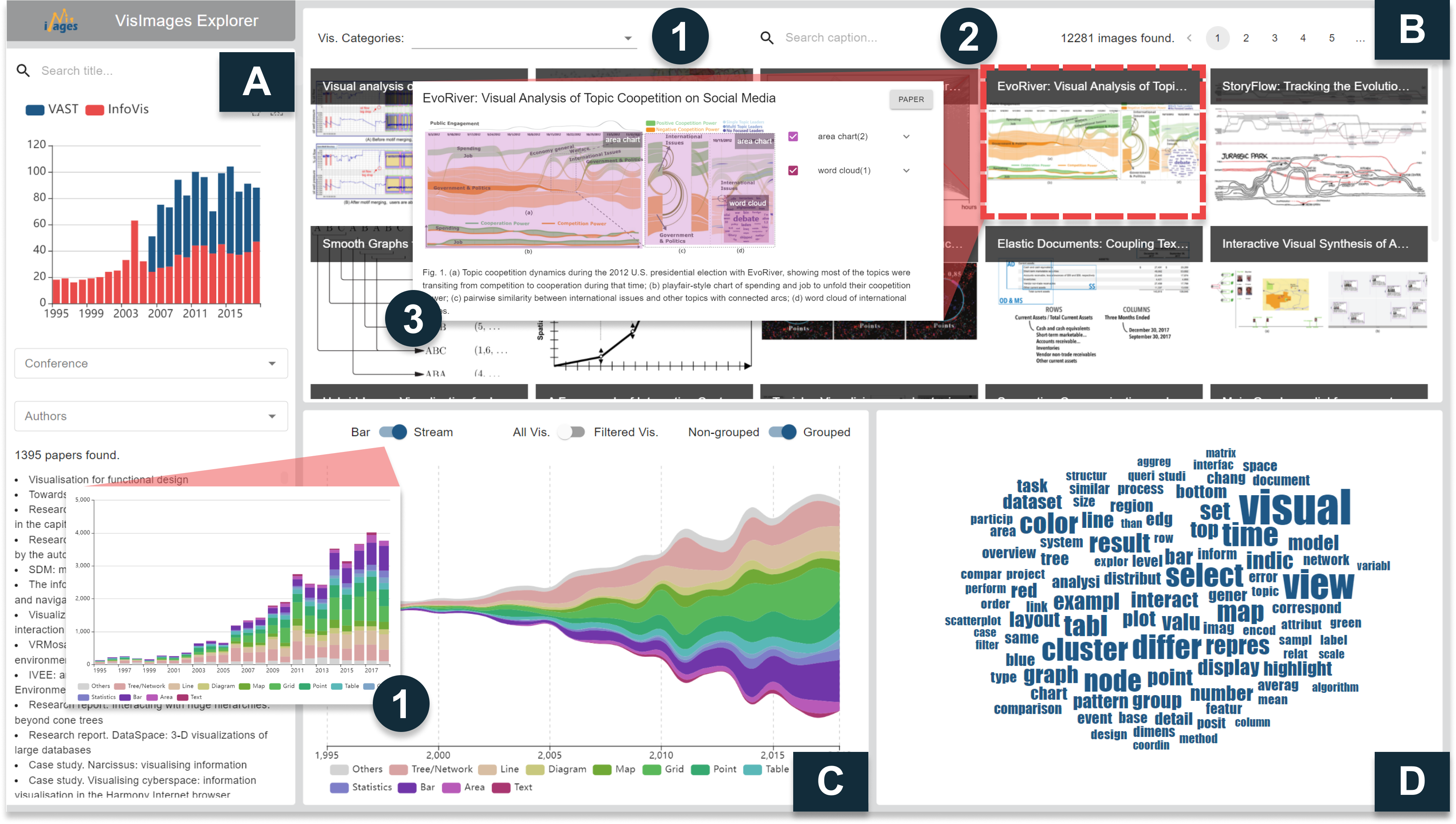}
    \caption{VisImages Explorer. (A) is a paper search panel filtering papers by title, year, conference, and author. (B) is an image gallery view displaying the images in the filtered papers. The gallery is facilitated with a visualization selector (B1) and a caption searcher (B2) for further filtering of the images. Users can click the images of interest to view the detailed annotations (B3). (C) is the visualization distribution view showing the numbers of visualizations. Users can switch between a stream graph and a bar chart for different preferences (C1). (D) is a word cloud view exhibiting the word frequencies in the captions.}
    \label{fig:interface}
\end{figure}
\subsection{Investigating the Use of Visualizations}
\label{sec:explorer}
VisImages contains rich information from IEEE VIS publications, including images, visualization types and their bounding boxes, captions, and publication metadata.
The information can be used to understand the visual designs and the papers, which might be useful for junior visualization researchers and novice visualization designers.
Therefore, to assist these users in data exploration, we develop \textbf{VisImages Explorer} for efficient data filtering and searching.

VisImages Explorer consists of a paper search panel, an image gallery view, a visualization distribution view, and a word cloud view.
The \textbf{paper search panel} (\figurename\ref{fig:interface}-A) allows users to filter papers by titles, years, conferences, and authors.
A histogram is displayed to show the number of publications from different conferences across years.
The \textbf{image gallery view} (\figurename\ref{fig:interface}-B) then exhibits all images based on the paper searching results. 
Users are allowed to further filter the images by visualization types (\figurename\ref{fig:interface}-B1) and keywords in the captions.
Users can examine the annotations of each image in a detailed view by clicking the image.
Moreover, the explorer has a \textbf{visualization distribution view} (\figurename\ref{fig:interface}-C) showing the number of visualizations and a \textbf{word cloud view} (\figurename\ref{fig:interface}-D) visualizing the word frequencies in the captions of the filtered images.
To demonstrate the usefulness of VisImages, we present two scenarios for junior visualization researchers and novice visualization designers.

\textbf{Scenario 1: Reviewing papers of graph visualizations.}
Suppose James, a junior PhD student, is surveying papers about graph visualizations in IEEE VIS to understand the field.
Papers might not explicitly mention ``graph'' in the title, so he decides to search the images directly.
He turns to the image gallery view and filters the images whose captions contain the keyword of ``graph'' (\figurename\ref{fig:interface-case}-A2).
For the reason that graph data can be represented with a matrix or node-link diagram, he first filters the images that contain the sub-type of ``matrix,'' but only retrieves 58 images.
Then he filters the images with the sub-type of ``node-link diagram'' (denoted as ``graph'' in this work, as shown in \figurename\ref{fig:interface-case}-A1) and retrieved 385 images, which means that node-link diagram might be a much more popular representation type for graph data.
Due to different research targets of VAST and InfoVis, James further investigates the images of two conferences respectively.
He discovers that the top words in the word cloud views of two conferences are different.
In addition to the word ``graph'' and ``visual,'' the top words of InfoVis are ``layout'' and ``edge'' (\figurename\ref{fig:interface-case}-A3) while the top words of VAST are ``view'' and ``cluster'' (\figurename\ref{fig:interface-case}-A6).

When exploring the captions of InfoVis in depth, James discovers that the images are mostly about the resulted graph visualizations after applying layout optimization algorithms (\figurename\ref{fig:interface-case}-A4).
Clicking on the word ``layout'' in the word cloud view, the images are further filtered if their captions contain the word.
He immediately discovers that the titles of these papers contain keywords describing the algorithms, such as ``topological fisheye,'' ``stress majorization,'' and ``divided edge bundling'' (\figurename\ref{fig:interface-case}-A5).
Therefore, he obtains a paper collection about graph layout optimization for further reading, which is a good starting point for his research career.

James further explores the images in VAST, where the word ``view'' and ``cluster'' frequently occur in the captions. Different from InfoVis, graph visualizations in VAST images usually appear in the views of visual analytics (VA) systems. As a result, the words ``view'' commonly co-occurs with the word ``graph'' in the captions. Besides, the ``cluster'' patterns of graphs are the patterns described the most in the VA systems with graphs. He then clicks on the word ``cluster'' in the word cloud view and investigates the captions. From the captions, he understands that clusters imply patterns of interest and help to guide users for further exploration.

\textbf{Scenario 2: Inspiring visual designs.} Suppose Mary, a junior visualization designer, is designing a map visualization for geospatial data.
She wonders if there are any design strategies to improve the map visualization to represent more data attributes.
She first filters the images that contain a map. The stream graph in the visualization distribution view indicates that heatmaps and bar charts are the most common sub-types that co-occur with maps (\figurename\ref{fig:interface-case}-B).
It is intuitive that heatmaps can visualize density data on maps, but Mary has no clear idea how bar charts can display together with maps.
Therefore, she additionally selects bar charts, and the images containing both maps and bar charts are filtered.
The histogram shows the distribution of bar charts and maps in the filtered images (\figurename\ref{fig:interface-case}-C).
By investigating the images, she discovers that bar charts and maps can be organized without overlapping, such as positioning them side-by-side (\figurename\ref{fig:interface-case}-D1) or surrounding the map with circular bar charts (\figurename\ref{fig:interface-case}-D2). 
When there is overlapping, the bar charts can be tooltips floating on the map (\figurename\ref{fig:interface-case}-D3) or glyphs positioned at the places of interest (\figurename\ref{fig:interface-case}-D4).
Mary is inspired by the designs in the retrieved images and decides on a suitable one depending on design requirements.

\subsection{Classification Benchmarking with VisImages}
\label{sec:classification}
Object classification has been adopted in many visualization scenarios, such as visualization reverse-engineering~\cite{poco2017reverse, zhou2021reverse, ying2021glyphcreator}, visualization demographic analysis~\cite{battle2018beagle}, and chart data extraction~\cite{jung2017chartsense}.
In this case, we show how VisImages can serve as a benchmark for visualization classification models using the annotations of visualization sub-types.

\subsubsection{Experiment Setup}
To compare VisImages and other datasets in training classification models, we set up experiments to mutually evaluate the models trained on different datasets.
Specifically, we train the models on one dataset and evaluate them on another, and investigate their performance in different situations.
We select Beagle~\cite{battle2018beagle} as a baseline dataset to train classification models because Beagle has the most classes in common with VisImages and the largest sample number and class number among existing datasets.

\textbf{Data Processing.} 
We select 17 common classes from VisImages and Beagle.
We convert the images in Beagle into bitmap images for model training and evaluation since they are in the SVG format originally.
For the images in VisImages, we crop the visualizations from the images and categorized them by sub-types (denoted as VisImages-cropped).
Before the experiment, we randomly divide Beagle and VisImages into training (75\%) and test (25\%) sets.

\textbf{Models.} In the experiments, we select two widely-used object classification models with different numbers of layers, i.e., ResNet-50, ResNet-101~\cite{he2016deep}, VGG-16, and VGG-19~\cite{simonyan2014very}. Taking an image as input, the models will output the probabilities of specific classes the image belongs to.

\textbf{Training.} We follow a similar training process described by Krizhevsky et al.~\cite{krizhevsky2012imagenet} that all models are trained in two stages with weights pre-trained on ImageNet~\cite{deng2009imagenet}.
In the first stage, we freeze the weights of convolutional layers and train the classification heads.
In the second stage, we unfreeze the convolutional layers and finetune the overall weights.
In each stage, the models are trained with stochastic gradient descent (SGD) with $1e^{5}$ steps.
The initial learning rates of ResNets and VGGNets are $1e^{-3}$ and $1e^{-5}$, respectively. 
We use categorical cross-entropy as loss function.
Because of the imbalanced distribution between different classes, we introduce class weights for loss computation to reduce over-fitting on specific classes.

\textbf{Metrics.} Following the standard evaluation protocol of multi-classification problems~\cite{he2016deep}, we use the top-k accuracy as a metric, which means, if the ground-truth label appears in the top k predictions, the classification is regarded correct.
The top-1 and top-3 accuracies are shown in \tablename~\ref{tab:recognition}.

\begin{table}[tb]
    \centering
    \scriptsize%
    \caption{The top-1 and top-3 accuracies (\%) of different models on visualization classification under different situations. The underlined numbers denote the highest accuracies among the models.}
    \begin{tabular}{lcccc}
    \toprule
    Training Set &\multicolumn{2}{c}{(A) Beagles}&\multicolumn{2}{c}{(B) VisImages}\\
    Test Set & Beagle  & VisImages (Acc$\downarrow$)& VisImages & Beagle (Acc$\downarrow$) \\\midrule \midrule
    ResNet-50 & 79.3/\underline{99.0} & 32.6/38.8&  78.3/92.6 & \textbf{10.6/9.5}  \\\midrule
    ResNet-101 & \underline{80.6}/98.9 &26.0/33.2& \underline{78.9}/\underline{94.4} & \textbf{11.5/9.4} \\\midrule
    VGG-16 &79.9/98.7 & 36.7/43.8&77.5/92.7 & \textbf{8.7/7.1}\\\midrule
    VGG-19 &80.1/98.7 &34.8/40.4&76.9/92.2 & \textbf{8.3/6.0} \\
    \bottomrule
    \end{tabular}
    \label{tab:recognition}
\end{table}

\begin{figure*}[!htb]
    \centering
    \includegraphics[width=\linewidth]{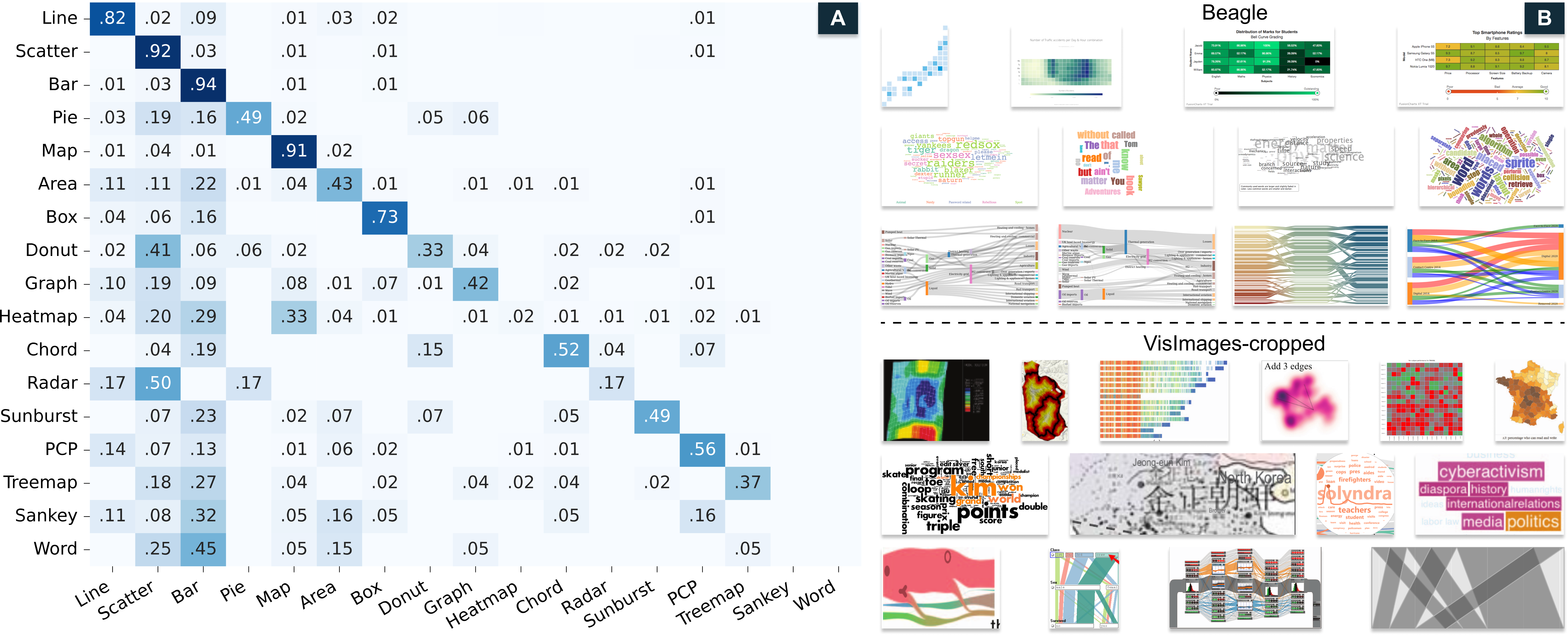}
    \caption{(A) Confusion matrix of ResNet-101 trained from Beagle and tested on VisImages (on the left). Rows of matrix represent the ground truth, and columns represent the predictions. The confusion matrix is normalized by rows. (B) Examples of heatmap, word cloud, and Sankey diagram from Beagle and VisImages-cropped, respectively.}
    % \vspace{-10pt}
    \label{fig:confusion}
\end{figure*}

\subsubsection{Classification Performance Analysis}
We analyze the model performance from different aspects.

\textbf{CNN Model Comparison.} We first observe the performance of different models, and discover that ResNet-101 achieves the best performance on VisImages and Beagle during training (underlined values in the second row).
The results conform to the conclusion drawn by He et al.~\cite{he2016deep} that deeper architectures of ResNets make the models achieve better results than VGGNets in object classification.

Although CNNs achieve satisfactory performance on classifying bitmap visualizations, there might be some setbacks of the models when compared to other methods.
Battle et al.~\cite{battle2018beagle} classified the visualizations in Beagle with decision trees and achieved 86.5\% top-1 accuracy, which is better than the performance of CNNs.
Differently, the decision trees take in hand-crafted features of SVG visualizations, instead of visual features extracted from bitmap images.
Specifically, the features include style features (e.g., number of fill and border colors and stroke width) and per-element features (e.g., CSS class names, circle elements, and rect elements).
These features might provide semantic information about the visualizations that are useful for visualization classification.
Therefore, there might be some limitations of CNNs for visualization classification, which conforms to the insights provided by Haehn et al.~\cite{haehn2019cnn} that ``\textit{CNN architecture performance on natural images is not a good predictor for performance on graphical perception tasks}.''

\textbf{Training Dataset Comparison.} Training data is critical to model generalizability. 
Overall, we discover that models trained on Beagle have higher accuracies than the models trained on VisImages. However, the comparison is not fair because the test sets are not the same.
Therefore, we conduct cross-evaluation by training the models on one dataset but testing them on the other one.
First, we evaluate the models trained on Beagle with VisImages.
The models encounter a steep decrease on both top-1 (26\%-34.8\%) and top-3 (33.2\%-40.4\%) accuracies when tested on VisImages (\tablename~\ref{tab:recognition}(A)).
Second, we evaluate the models trained on VisImages on Beagle and discover that models also encounter a decrease on the top-1 (8.3\%-11.5\%) and top-3 (6.0\%-9.5\%) accuracies (\tablename~\ref{tab:recognition}(B)). However, the decrease is much smaller compared to the previous ones.
The model performance might decrease when testing the model on a dataset with a different ``data generating process'' because of the generalization error~\cite{bengio2017deep}.
The inhibition of model performance decrease indicates that models trained on VisImages might have better generalizability compared to Beagle.

\textbf{Confusion Analysis.} To further investigate the reason for the performance decrease of the models trained on Beagle when testing on VisImages, we visualize the confusion matrix of ResNet-101, which achieves the best performance (\figurename\ref{fig:confusion}, left).
Rows of the matrix encode the ground-truth labels and columns encode the prediction labels. We arrange the row and column labels by the sample number of classes in Beagle decreasingly. The matrix is normalized row-wise, and thus the cells on the diagonal represent the recall rates.

In the matrix, while most cells on the diagonal are in dark blue, there are some cells in light colors, such as heatmap, Sankey diagram, and word cloud, indicating low recall rates for these classes. By randomly selecting the samples of these classes from both datasets, we discover that the visualizations of these classes in VisImages are more diverse in appearance, such as layout, color, and shape, but the ones in Beagle are similar in design, as shown in \figurename\ref{fig:confusion}, right.
The reason might be that the charts in Beagle are generated by similar visualization libraries (e.g., D3 and Plotly) with default settings of styles.
However, in VisImages, many of charts in VisImages are created from scratch using design tools (e.g., Adobe Illustrator) or low-level programming languages (e.g., Javascript).
Due to the higher diversity in layout and design of the samples, VisImages can be a good benchmark and complementary for existing datasets.

In addition, we discover that there are blue clusters in the bottom-left of the confusion matrix, especially the first three columns. The clusters indicate that the model usually misclassifies the visualizations to be line chart, scatterplot, and bar chart. We infer that the confusion might be caused by the large numbers of samples of these three classes (about 90\%), even though we have introduced class weights to relieve the over-fitting on these classes. Besides, many of the test samples in VisImages might have different styles that are not ``seen'' by the models during training. Therefore, when these samples occur, the model might tend to ``guess'' their labels with popular classes. To reduce the confusion, balancing the class distribution and adding more samples to smaller classes might be a practical solution, such as combining Beagle and VisImages as a training set.

\subsection{Visualization Localization with VisImages}
\label{sec:localization}
While VIS30K~\cite{chen2021vis30k} and Viziometrics~\cite{lee2017viziometrics} only focus on the classification of the image usage in the papers (e.g., equation, diagram, photo, and plot), VisImages further specifies the positions (i.e., bounding boxes) of visualizations in the images.
In this case, we exhibit the use of visualization bounding boxes, a featured dimension of VisImages dataset that can be used to train object localization models.
Specifically, the models can be used to localize visualizations from the images for reverse engineering of visual analytics (VA) systems and visualization position analysis.

\subsubsection{Training Visualization Localization Models}
We first train Faster R-CNN~\cite{ren2015faster}, one of the most widely-used object localization models, to predict the position of visualizations in the images.
We used 80\% of images for training and 20\% for testing.
Following a similar pipeline of Ren et al.~\cite{ren2015faster}, we train the model using SGD optimizer with a learning rate of $1e^{-3}$ for 15k mini-batches.
A momentum of 0.9 and a weight decay of $1e^{-4}$ are adopted.

\begin{figure*}[!tb]
    \centering
    \includegraphics[width=\linewidth]{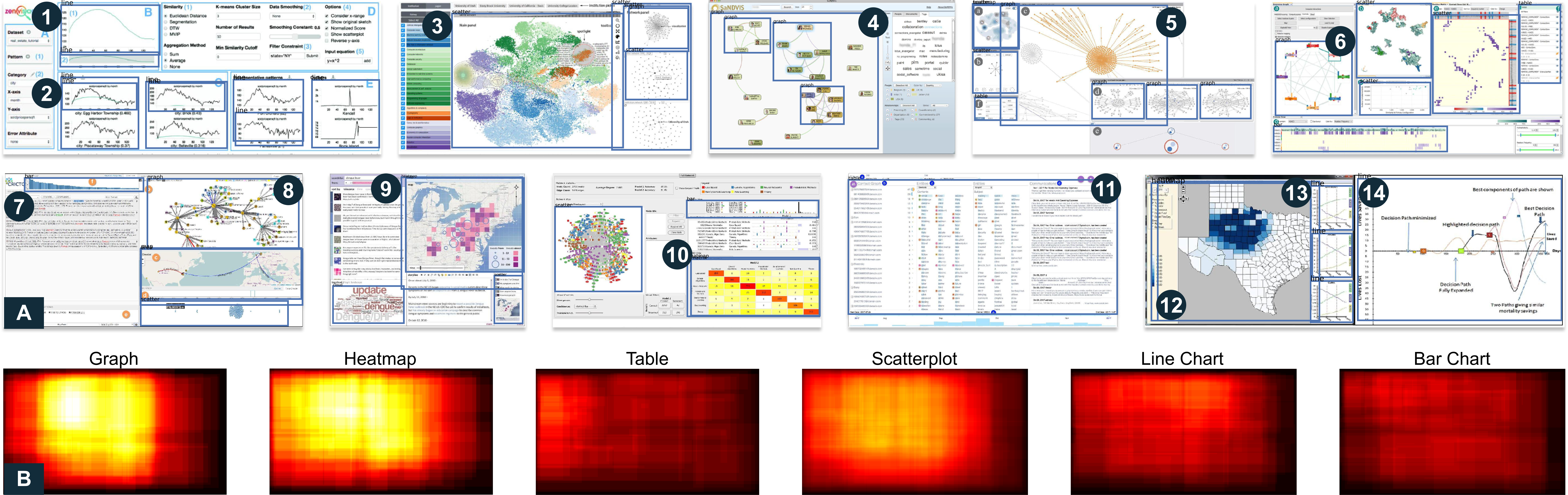}
    \caption{(A) Results of visualization localization in visual analytics systems and (B) heatmaps showing the spatial distributions of the visualizations.}
\label{fig:va_heatmap}
\end{figure*}

\begin{table}
    \scriptsize
    \caption{APs under different IoU thresholds.}
    \begin{tabular}{lccc}
    \toprule
    \textbf{Sub-type}    & $AP_{IoU=0.50}$ & $AP_{IoU=0.75}$ & $AP_{IoU=0.90}$ \\\midrule
    graph                & \textbf{0.96}   & \textbf{0.96}   & \textbf{0.70}   \\
    table                & 0.95   & 0.90   & 0.69   \\
    scatterplot          & 0.83   & 0.77   & 0.29   \\
    line chart           & 0.82   & 0.71   & 0.23   \\
    heatmap              & 0.80   & 0.80   & 0.40   \\
    flow diagram         & 0.75   & 0.70   & 0.46   \\
    bar chart            & 0.69   & 0.58   & 0.10   \\
    map                  & 0.68   & 0.64   & 0.54   \\
    parallel coordinate  & 0.67   & 0.52   & 0.28   \\\midrule
    $mAP$                  & 0.78   & 0.78   & 0.52   \\\bottomrule
    \end{tabular}
\label{tab:map}
\end{table}

We use average precision (AP) to evaluate the model performance on object detection~\cite{Everingham15}.
Besides, the IoU is used to measure the overlap between a predicted box and a ground-truth box.
For a detailed definition of IoU, please refer to Section~\ref{sec:iou} and \figurename\ref{fig:metric}-A.
\tablename~\ref{tab:map} shows the APs of different visualization sub-types, as well as mean average precision (mAP) under different IoU thresholds.

\subsubsection{Model Inference on VA Systems}
After training visualization localization models, we further apply the models on VA system interfaces.
We use the images from the Multiple-View Visualization (MV) dataset~\cite{chen2020composition}, which contains 360 VA interfaces cropped from the publication figures.
Samples of inference results are exhibited in \figurename\ref{fig:va_heatmap}.
Overall, we discover that Faster-RCNN trained with VisImages can successfully localize different visualizations (or views) in the interface.
In addition, the model can also handle various challenging cases: 1) visualizations with sub-structures, e.g., graph with sub-graphs (\figurename\ref{fig:va_heatmap}-A4); 2) visualizations with composite sub-types, e.g., heatmap + map (\figurename\ref{fig:va_heatmap}-A9, A13) and heatmap + matrix (\figurename\ref{fig:va_heatmap}-A10); and 3) visualizations with extreme aspect ratios (\figurename\ref{fig:va_heatmap}-A7).
The results demonstrate the potential of using computer vision models for reverse engineering the VA system designs in the publications, by considering the data flow and interactions among views~\cite{chen2021nebula, su2021natural, tominski2021toward}.

\subsubsection{Localization Performance Analysis}
\label{sec:failed-cases}
We analyze the localization performance of the model combining the APs and samples of VA systems.

Overall, the mAPs on VisImages are 0.78 (IoU=0.50), 0.78 (IoU=0.75), and 0.52 (IoU=0.90).
The graph achieves the highest AP (IoU=0.50) of 0.96, followed by table, scatterplot, and line chart.
Interestingly, even though bar chart has the largest portions of samples, its AP is only ranked the $7^{th}$ (IoU=0.50).
When investigating in detail, the APs of basic charts (i.e., bar chart, line chart, and scatterplot, which are commonly facilitated with the coordinate systems) decrease drastically with the rise of IoU threshold, indicating that precisely determining the bounding box of these visualizations are challenging for the model.
For example, when the same types of visualizations are aligned closely (\figurename\ref{fig:va_heatmap}-A2), the model will fail to determine the area of the visualizations.
By exploring the samples, we discover that these sub-types have diverse configurations (with or without the coordinate axes) and sizes, possibly making the model confused about the precise borders for these visualizations.
Differently, graph (\figurename\ref{fig:va_heatmap}-A4, A5, A6, A8), table (\figurename\ref{fig:va_heatmap}-A11), and map (\figurename\ref{fig:va_heatmap}-A9, A13) usually occupy a large area in the images, which might make the localization of these sub-types relatively simpler for the model.

\subsubsection{Spatial Distribution of Visualizations in VA Systems}
The model inference on VA systems can help us understand how different visualization types are distributed spatially.
Although the inference might not be perfectly correct, we argue that the results can indicate general distributions of the visualizations because of the acceptable APs shown in \tablename~\ref{tab:map}.
We visualize the spatial distribution of specific types of visualization using heatmaps.
To plot a heatmap, we transform all VA system images to the same scale, derive the transformed bounding boxes of the visualizations, and overlay all bounding boxes on the same canvas.
The heatmap is normalized by the total number of the bounding boxes of this type.
To facilitate comparison, we use a consistent brightness scale for different visualization types.
In \figurename\ref{fig:va_heatmap}-B, we visualize the heatmaps of the most popular visualizations in the VA samples.

From the figure, we discover that some heatmaps have extremely bright areas, i.e., the graph and heatmap visualizations.
Therefore, graph and heatmap visualizations have a higher density of visualizations in the bright area compared to other visualizations.
In addition, these visualizations are commonly distributed in the top-left of the VA systems, usually as the main view of the system.
On the contrary, the table, scatterplot, line chart, and bar chart visualizations do not have an extreme concentration, but these visualizations also have specific distribution patterns.
For example, table visualizations appear more on the left side, possibly serving as an auxiliary view for reference of raw data; line charts are more distributed on the top; scatterplots are also distributed more on the top-left; bar charts, the most commonly used visualizations, exhibit a relatively even distribution in the interface.
\section{discussion}
We see VisImages as an exciting start point for leveraging the intelligence of the visualization community itself and forge a path to a high-quality, fine-grained, and large-scale visualization dataset. 
We envision that VisImages can inform opportunities for advancing our knowledge of the field and the research of AI4VIS~\cite{wu2021survey}.

\textbf{Benefits to literature analysis.}
VisImages offers new possibilities to conduct literature analysis in visualization and help understand the evolution of the field. 
For example, VisImages Explorer (Section~\ref{sec:explorer}) shows the potential to help users discover the papers of interest and inspire design ideas for novice researchers and designers.
We make the explorer available for the community to discover more insights. 
Furthermore, researchers can investigate VisImages combined with existing publication metadata collections (e.g., keyvis~\cite{isenberg2016keyvis}), which is supposed to reveal new insights.
For example, analyzing the visual design patterns under different domains and topics.

\textbf{Opportunities for AI4VIS.}
VisImages can serve as a useful AI4VIS dataset~\cite{wu2021survey}, such as visualization classification (Section~\ref{sec:classification}), localization (Section~\ref{sec:localization}), visualization-text translation, and recommendation.
Visualization-text translation aims to construct relations between textual descriptions and visualizations. 
Images with elaborated captions in VisImages can naturally be a qualified resource for training such models for the scenarios like visual storytelling~\cite{shu2021dancingwords, shu2020makes, zhang2021visual}.
Moreover, a line of research puts efforts on visualization recommendation~\cite{li2021kg4vis, wu2021multivision}, for example, suggesting potential layouts and combinations of different visualization types in a design~\cite{chen2020composition}.
VisImages contains images of well-crafted VA systems for different topics, such as sports analysis~\cite{wang2021tac} and urban planning~\cite{deng2021compass}.
Along with annotations of visualization types and positions, the dataset provides a valuable resource for training models on such tasks.

\textbf{Limitations.}
Despite the significance and usefulness of VisImages, it still has limitations.
First, we tried best to ensure the quality of our annotation with a series of measures, such as the gold standards, majority voting, and sampling test.
Mislabeling is inevitable, especially in the situation where recognizing visualization and their variations requires significant expertise. 
As an alternative, we greatly welcome the visualization community, especially the authors of the publications, to examine and possibly correct the mislabeled visualizations.
Second, our dataset currently contains three major types of labels (i.e., image captions, visualization types, and bounding boxes), leaving a wealth of information unexplored, such as axis titles and marks.
With richer information, VisImages can be used in a wider range of applications and support more complex tasks.
Third, our exemplar use cases only demonstrate featured usage of VisImages.
For example, in Section~\ref{sec:classification}, we first compare VisImages with Beagle, as Beagle has the most classes in common and the largest samples. 
Further comparison with more datasets (e.g., ChartSense~\cite{jung2017chartsense} and VizioMetrics~\cite{lee2017viziometrics}) may also benefit the analysis of low-level and high-level features, as well as the model performance. 
The exploration on extensive datasets is beyond the scope of the paper, but our use cases provide pointers to designing experiments and conducting analysis for future work. 
A more in-depth investigation of VisImages can offer illuminating insights and reach broad impacts.
\section{conclusion and future work}
In this paper, we create and make available VisImages, a visualization dataset from the top-venue visualization publications. 
VisImages includes 12,267 images with captions from 1,397 papers of IEEE InfoVis and VAST.
Each image is annotated with visualization types and their positions in the image, resulting in a total of 35,096 bounding boxes.
We further investigate VisImages with an overview of visualization distribution across years and types.
Besides, VisImages presents a more balanced distribution in the visualization types, compared to other state-of-the-art datasets in the visualization field.
The usefulness and significance of VisImages are demonstrated through three use cases, including visual literature review, visualization classification, and visualization localization.
We envision that VisImages can broaden the diversity of visualization research~\cite{lee2019broadening} and inspire new research opportunities.

However, VisImages only takes a first step to explore images in the visualization publications.
In the future, we intend to expand VisImages to cover more images from other top-notch journals and conferences, such as TVCG, CHI, and EuroVis.
Second, given the increasing number of images, we plan to develop a semi-automatic annotation method that leverages human and machine intelligence~\cite{deng2021eventanchor}. 
Third, we plan to gradually refine and improve the taxonomy to meet the growing diversity of visualization designs.

% or


% \appendices

% \section{}

\ifCLASSOPTIONcompsoc
  \section*{Acknowledgments}
\else
  \section*{Acknowledgment}
\fi

This work was supported by NSFC (62072400, 62002331) and the Collaborative Innovation Center of Artificial Intelligence by MOE and Zhejiang Provincial Government (ZJU).
This work was also partially funded by Zhejiang Lab (2020KE0AA02, 2021KE0AC02) and Microsoft Research Asia.
Our deepest gratitude went to the anonymous reviewers for their valuable comments that helped us improve this paper substantially.
We sincerely thank the researchers and the students from Zhejiang University and Zhejiang Lab for their time and effort in data annotation. 
Specifically, We thank Mengye Xu for her contributions in designing the VisImages logo and documenting the training materials for data annotation.

% \ifCLASSOPTIONcaptionsoff
%   \newpage
% \fi



\bibliographystyle{IEEEtran}

\bibliography{main.bib}


% \end{IEEEbiography}


\begin{IEEEbiography}[{\includegraphics[width=1in,height=1.25in,clip,keepaspectratio]{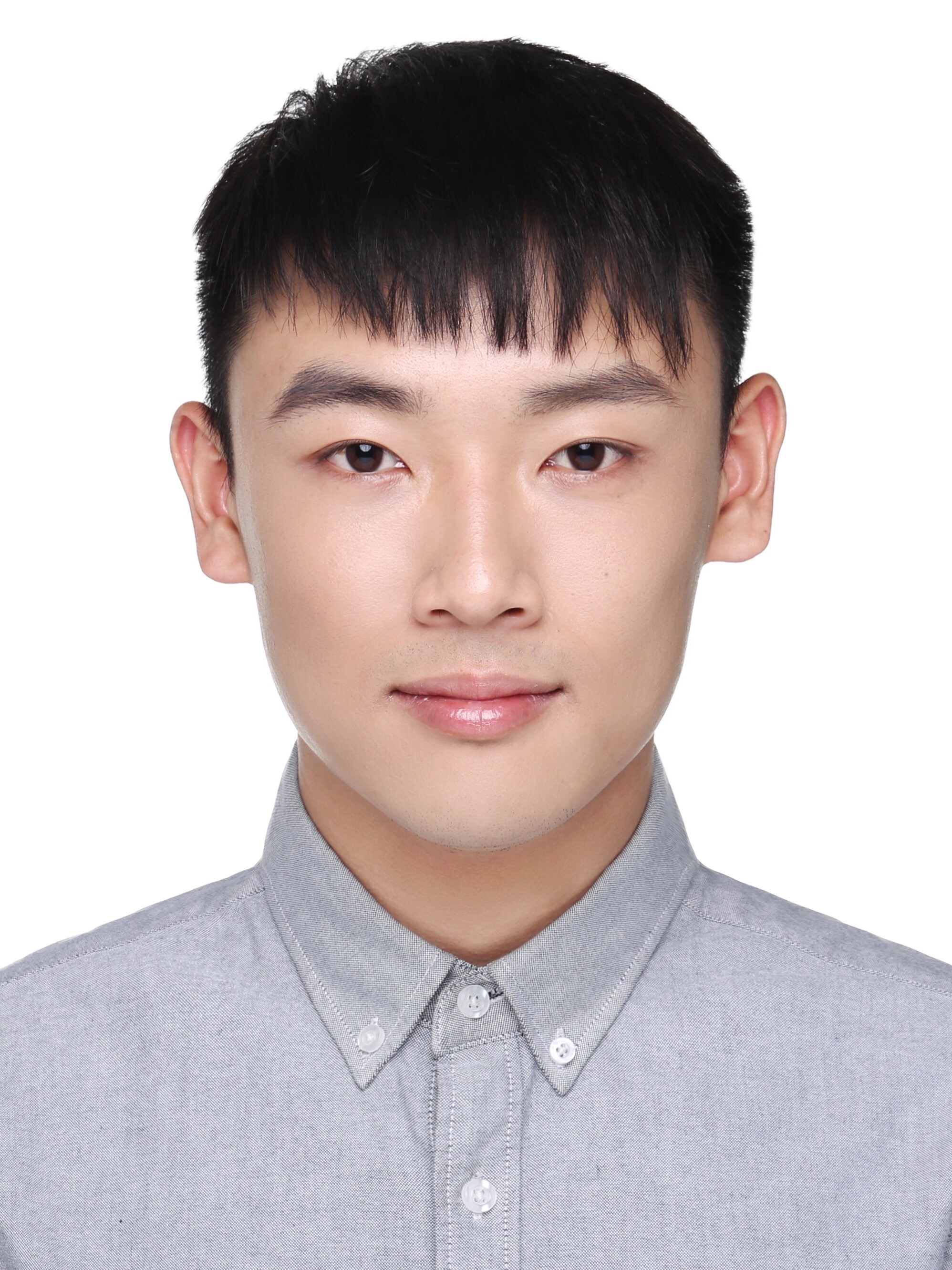}}]{Dazhen Deng}
is currently a Ph.D. student in the State Key Lab of CAD\&CG, Zhejiang University.
He received the B.E. degree in Applied Mathematics from Zhejiang University in 2018.
His research interests mainly lie in sports visualization and machine learning for visual analytics. For more information, please visit https://dazhendeng.github.io/.
\end{IEEEbiography}

\begin{IEEEbiography}[{\includegraphics[width=1in,height=1.25in,clip,keepaspectratio]{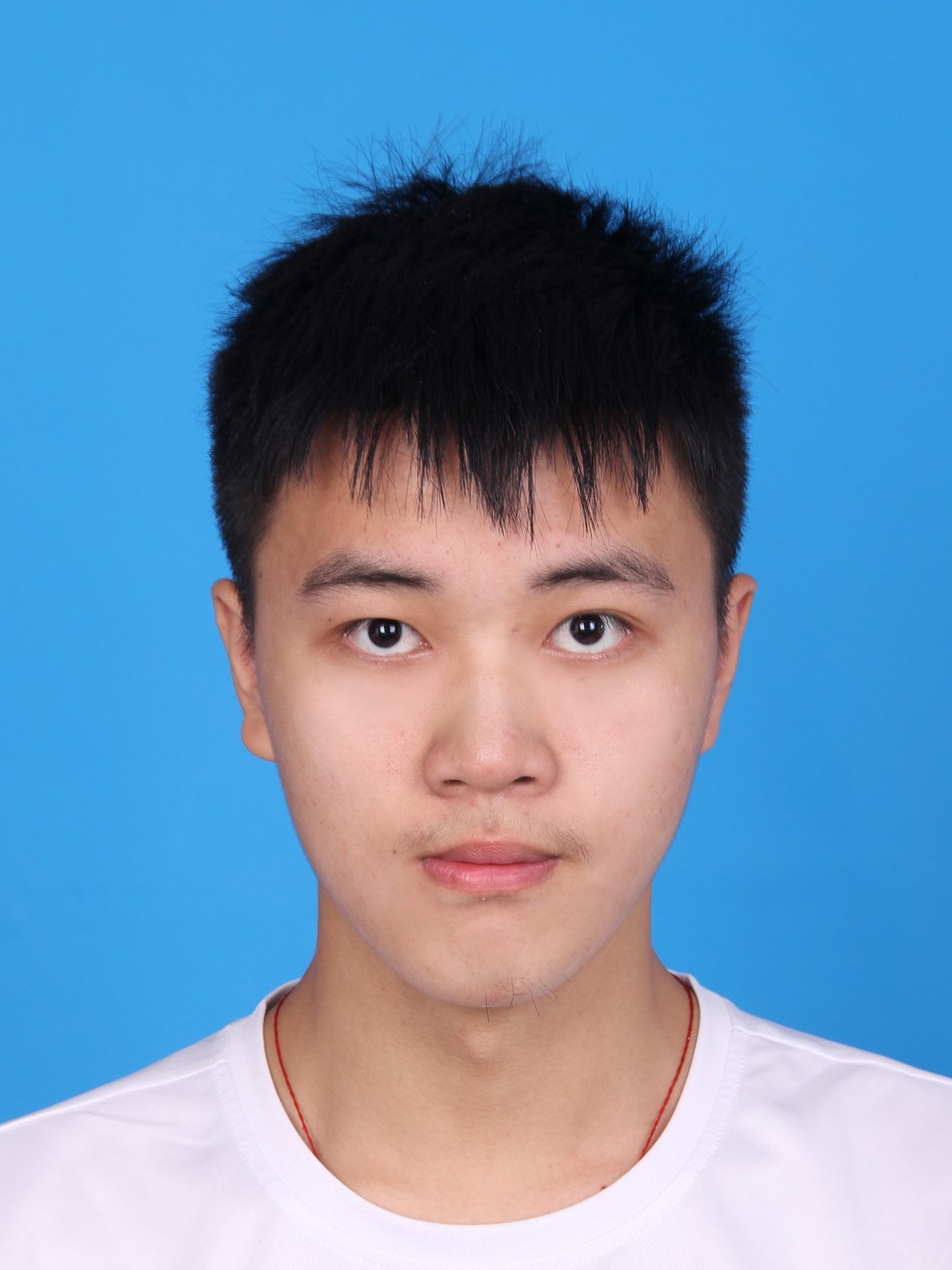}}]{Yihong Wu}
is currently a Ph.D. student in the State Key Lab of CAD\&CG, Zhejiang University.
He received the B.E. degree from Zhejiang University in 2020.
His research interests mainly lie in computer vision and visual analytics.
\end{IEEEbiography}

\begin{IEEEbiography}[{\includegraphics[width=1in,height=1.25in,clip,keepaspectratio]{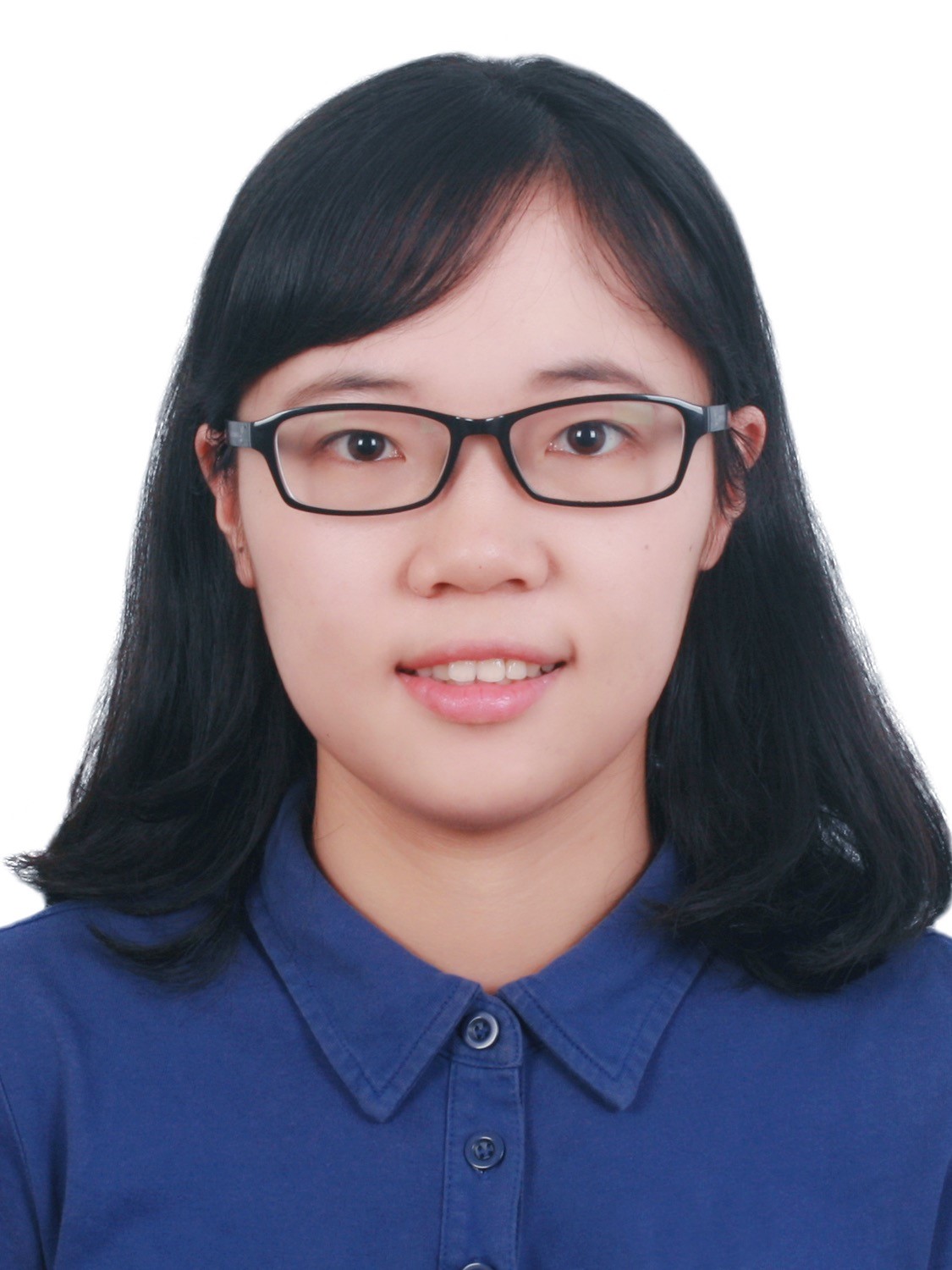}}]{Xinhuan Shu}
is currently a postdoctoral researcher in the Department of Computer Science and Engineering at the Hong Kong University of Science and Technology (HKUST). 
She received her Ph.D. degree from HKUST and her B.E. degree in Computer Science and Technology from Zhejiang University, China. 
Her research interests include visual data communication, animated visualization, and visual analytics. 
\end{IEEEbiography}

\begin{IEEEbiography}[{\includegraphics[width=1in,height=1.25in,clip,keepaspectratio]{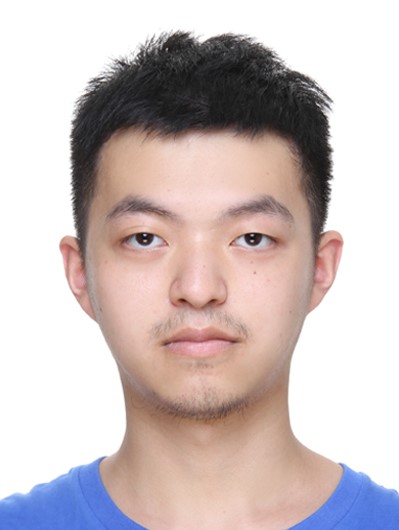}}]{Jiang Wu}
is currently a Ph.D. student in the State Key Lab of CAD\&CG, Zhejiang University.
He received the B.E. degree in Computer Science and Technology from Zhejiang University, China.
His research interests mainly lie in sports visualizations and event sequence analysis.
\end{IEEEbiography}

% \begin{IEEEbiography}{Mengye Xu}
% is currently a product manager at ByteDance Ltd. She received her B.S. degree in Digital Media Technology from Zhejiang University in 2021.
% \end{IEEEbiography}

\begin{IEEEbiography}[{\includegraphics[width=1in,height=1.25in,clip,keepaspectratio]{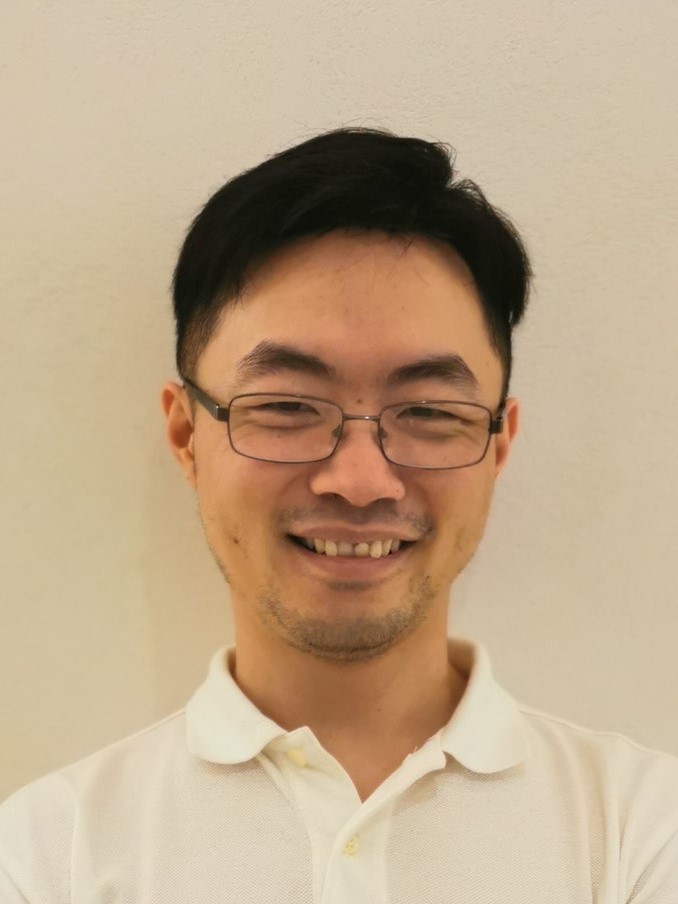}}]{Siwei Fu}
is an associate research scientist in Zhejiang Lab. 
His main research interests include visual analytics, intelligent user interface, and natural language interface. 
He received his Ph.D. degree in Computer Science and Engineering from the Hong Kong University of Science and Technology.
For more information, please visit https://fusiwei339.bitbucket.io/
\end{IEEEbiography}

\begin{IEEEbiography}[{\includegraphics[width=1in,height=1.25in,clip,keepaspectratio]{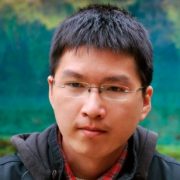}}]{Weiwei Cui}
  is a Principal Researcher at Microsoft Research Asia, China. He received his Ph.D. in Computer Science and Engineering from the Hong Kong University of Science and Technology and his B.S. in Computer Science and Technology from Tsinghua University, China. His primary research interests lie in visualization, with focuses on text, graph, and social media. For more information, please visit http://research.microsoft.com/en-us/um/people/weiweicu/
\end{IEEEbiography}

\begin{IEEEbiography}[{\includegraphics[width=1in,height=1.25in,clip,keepaspectratio]{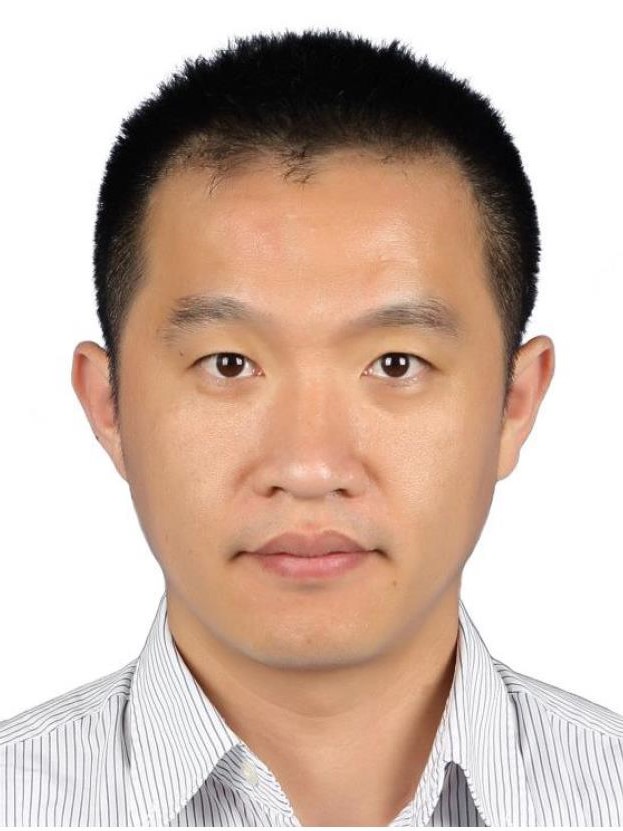}}]{Yingcai Wu}
  is a Professor at the State Key Lab of CAD\&CG, Zhejiang University. His main research interests are in information visualization and visual analytics, with focuses on sports science and urban computing. He received his Ph.D. degree in Computer Science from the Hong Kong University of Science and Technology. Prior to his current position, Dr. Wu was a postdoctoral researcher at the University of California, Davis from 2010 to 2012, and a researcher in Microsoft Research Asia from 2012 to 2015. For more information, please visit http://www.ycwu.org.
  \end{IEEEbiography}

% % insert where needed to balance the two columns on the last page with
% %\newpage



%\vfill


\end{document}